\documentclass[a4paper,fleqn]{cas-dc}

\usepackage[numbers,sort&compress]{natbib}

\usepackage[acronym,toc]{glossaries}
\newacronym{rl}{RL}{Reinforcement Learning}
\newacronym{cb}{CB}{Contextual Bandits}
\newacronym{ql}{QL}{Q-Learning}
\newacronym{dqn}{DQN}{Deep Q-Network}
\newacronym{rf}{RF}{Random Forest}
\newacronym{gb}{GB}{Gradient Boosting}
\newacronym{et}{ET}{Extra Trees}
\newacronym{hvac}{HVAC}{Heating, Ventilation, and Air Conditioning}
\usepackage[english]{babel}
\usepackage[nottoc]{tocbibind}
\usepackage{subcaption,graphicx}
\usepackage{lipsum}
\usepackage{adjustbox}
\usepackage{graphicx}
\usepackage{pgfplots}
\usepackage[ruled,vlined]{algorithm2e}
\usepackage{booktabs}
\usepackage{pdflscape}
\usepackage{orcidlink}
\usepackage{booktabs}
\usepackage{multirow}
\usepackage{siunitx}
\usepackage{pgfplots}
\usepackage{pgfplotstable}
\pgfplotsset{compat=1.18}
\usepackage{graphicx}
\usepackage{subcaption}
\usepackage{graphicx}
\usepackage{booktabs}
\usepackage{float}
\usepackage{tikz}
\usepackage{pgfplots}
\usepgfplotslibrary{groupplots}
\usepgfplotslibrary{statistics}
\usepackage{booktabs}
\usepackage{subcaption}
\usepackage{amsmath}
\usepackage{placeins}
\pgfplotsset{compat=1.18}
\usepackage{caption}
\usepackage{appendix}

\begin{document}
\let\WriteBookmarks\relax
\def\floatpagepagefraction{1}
\def\textpagefraction{.001}
%\justifying

\shorttitle{From Thermal Preference Prediction to Adaptive Climate Control}

\shortauthors{Ihianle et~al.}

\title [mode = title]{From Thermal Preference Prediction to Adaptive Thermal Intervention: A Reinforcement Learning Approach Using Physiological and Environmental Sensing}                      

\tnotetext[1]{This document is the results of the research
   project funded by the Nottingham Trent }

\author[1]{Isibor Kennedy Ihianle\orcidlink{0000-0001-7445-8573}}

\cormark[1]
\fnmark[]

\ead{isibor.ihianle@ntu.ac.uk}

%\credit{Maximising Relevance, Minimising Redundancy: HRV Feature Selection for Stress Classification}

\affiliation[1]{organization={Department of Computer Science, Nottingham Trent University},
    %addressline={Radarweg 29}, 
     city={Nottingham},
    % citysep={}, % Uncomment if no comma needed between city and postcode
    postcode={NG11 8NS}, 
    country={United Kingdom}}

\affiliation[2]{organization={York University},
    %addressline={Radarweg 29}, 
    city={Ontario},
    % citysep={}, % Uncomment if no comma needed between city and postcode
    postcode={M3J 1P3}, 
    country={Canada}}

\author[1]{Emmanuel Manu\orcidlink{0000-0000-000-0000}}
\fnmark[]
\ead{emmanuel.manu@ntu.ac.uk}
\author[1]{Ehsan Asnaashari\orcidlink{0000-0001-5552-9428}}
\fnmark[]
\ead{ehsan.asnaashari@ntu.ac.uk}
\author[2]{Mojgan Jadidi\orcidlink{0000-0002-2692-6157}}
\fnmark[]
\ead{mjadidi@yorku.ca}
\author[1]{Pedro Machado\orcidlink{0000-0003-1760-3871}}
\fnmark[]
\ead{pedro.machado@ntu.ac.uk}
\author[1]{Amrit Sagoo\orcidlink{0000-0000-0000-0000}}
\fnmark[]
\ead{amrit.sagoo@ntu.ac.uk}
\author[1]{Ahmad Lotfi\orcidlink{0000-0002-5139-6565}}
\fnmark[]
\ead{ahmad.lotfi.ac.uk}

\cortext[cor1]{Corresponding author}
\begin{abstract}
Personalised thermal comfort is essential for occupant wellbeing and for the development of more responsive building-control strategies, yet conventional \gls*{hvac} systems rely on static setpoints and population-level comfort models that fail to capture individual physiological variability. This paper presents a two-stage personalised thermal comfort approach integrating multimodal physiological and environmental sensing with reinforcement learning-based decision-making. In Stage~1, participant-specific Comfort Oracles are developed using ensemble learning models trained on wearable physiological signals, including heart rate, wrist and ankle skin temperature and body-proximity temperature, combined with environmental sensing. The oracles estimate the probability that an occupant prefers a cooler environment, no temperature change, or a warmer environment, thereby representing both the predicted preference and the model’s confidence. In Stage~2, the personalised oracles are embedded within three \gls*{rl} controllers - \gls*{cb}, \gls*{ql} and \gls*{dqn} to recommend equivalent environmental-temperature interventions that maximise predicted comfort while penalising unnecessary or unstable actions. Experiments conducted using a publicly available dataset show that personalised comfort prediction performs consistently across participants, with environmental descriptors providing complementary improvements over physiological features alone. Within the proxy control environment, the Comfort Oracles support distinct occupant-specific intervention policies: \gls*{cb} achieves the highest predicted comfort and mean reward, \gls*{ql} provides the highest reward and comfort per unit of proxy intervention, and \gls*{dqn} provides greater action flexibility but increased intervention variability. These results demonstrate the feasibility of combining personalised comfort inference with reinforcement learning for adaptive thermal intervention-policy generation and provide a foundation for future integration with physically validated occupant-centric \gls*{hvac} systems.
\end{abstract}

% \includegraphics{figs/grabs.pdf}

% Keywords
\begin{keywords}
Thermal comfort \sep Comfort Modelling \sep HVAC control \sep Smart Buildings \sep Wearable Sensors \sep Reinforcement learning \sep Human-in-the-Loop AI
\end{keywords}

\maketitle

\section{Introduction}
\label{sec:introduction}
Thermal comfort is a critical determinant of wellbeing, productivity and satisfaction in residential and workplace environments. Suboptimal thermal conditions have been linked to reduced cognitive performance, impaired sleep quality and increased occupant complaints \citep{lan2011quantitative, arif2016impact}. Despite its significance, most \gls*{hvac} systems still rely on static setpoints, rule‑based schedules or population‑average comfort models including PMV/PPD \citep{fanger1970thermal}. These approaches assume homogeneity in thermal responses, yet decades of empirical evidence show substantial inter-individual variability driven by physiology, behavioural adaptation, clothing, metabolic rate and environmental history \citep{humphreys2002validity, de1998developing, kim2018personal2}. As a result, conventional \gls*{hvac} control often fails to deliver personalised comfort, leading to energy waste and frequent manual thermostat overrides.

Recent advances in wearable sensing and ubiquitous environmental monitoring have enabled continuous measurement of physiological signals such as heart rate, skin temperature and activity level, combined with environmental variables including outdoor temperature, humidity, wind speed and solar radiation. These multimodal data streams provide a rich foundation for personalised thermal comfort modelling. Machine learning approaches, including ensemble methods - \gls*{rf}, \gls*{gb} and \gls*{et}, have demonstrated strong predictive performance with sensor data, achieving accuracies between 0.65 and 0.85 across diverse occupants \citep{liu2019personal, abdelrahman2022personal, kim2018personal}. Furthermore, personalised models consistently outperform population-level comfort predictors, confirming that thermal preference is highly individualised and cannot be reliably inferred from aggregate behavioural patterns \citep{zhang2010thermal}.

However, prediction alone is insufficient for intelligent comfort management. A classifier may estimate whether an occupant prefers cooler, warmer or unchanged conditions, but an additional decision-making mechanism is required to translate that estimate into a candidate thermal intervention. This gap has motivated growing interest in \gls*{rl} for building control. Our earlier work demonstrated the feasibility of using physiological responses and individual thermal preferences to automatically determine consensus temperature set-points for shared environments. However, that approach relied on predefined control logic rather than adaptive sequential decision-making. The present work extends this direction by integrating personalised comfort prediction with reinforcement learning to generate adaptive intervention policies \citep{ihianle2022towards}. \gls*{rl} offers an approach for sequential decision-making under uncertainty and has demonstrated improved energy efficiency and comfort maintenance compared to rule‑based and PID controllers \citep{wei2017deep, yu2019deep, sivamayil2023systematic}. Deep \gls*{rl} approaches including \gls*{dqn} \citep{mnih2015human} and actor-critic methods \citep{duan2016benchmarking} have shown promise in learning complex building-control policies from high-dimensional state representations. Yet most \gls*{rl}-based thermal comfort studies rely on population-level comfort models, simplified thermal sensation labels or simulated building environments, limiting their ability to adapt to individual occupants. Furthermore, existing \gls*{rl} controllers typically assume access to indoor air temperature or \gls*{hvac} setpoint data, information not available in many wearable‑sensor datasets.

To address these limitations, this paper proposes a two-stage personalised thermal comfort architecture that integrates occupant-specific thermal preference prediction with \gls*{rl}-based intervention-policy generation. Stage~1 develops a personalised Comfort Oracle for each participant using physiological and environmental sensing. The oracle outputs a probability distribution over three thermal preference states - cooler, no change and warmer providing a continuous representation of comfort confidence. Stage~2 embeds this oracle within \gls*{rl} controllers - \gls*{cb}, \gls*{ql} and \gls*{dqn} to recommend an equivalent environmental-temperature intervention designed to maximise predicted comfort while penalising unnecessary or unstable actions. In this study, these actions are evaluated as proxy control signals representing the direction and relative magnitude of prospective thermal intervention rather than direct thermostat commands. The architecture therefore extends thermal comfort modelling from passive prediction to adaptive, occupant-centred intervention-policy evaluation, aligning with emerging trends in personalised intelligent environments \citep{liu2019personal,zhao2014data}.

The contributions of this paper are fourfold:
\begin{enumerate}
    \item A personalised thermal comfort modelling pipeline that integrates multimodal physiological and environmental sensing with ensemble learning to construct participant-specific probabilistic Comfort Oracles capable of representing inter-individual thermal preference variability.

    \item An \gls*{rl} formulation for personalised thermal intervention recommendation that embeds probabilistic Comfort Oracle outputs directly within the controller state, enabling \gls*{rl} policies to optimise comfort confidence rather than relying solely on discrete preference labels.

    \item A closed-loop proxy evaluation environment that enables equivalent environmental-temperature intervention policies to be trained and compared in the absence of measured indoor temperature, thermostat setpoints, actuator states and HVAC power-consumption data.

    \item A comparative evaluation of \gls*{cb}, \gls*{ql} and \gls*{dqn} across multiple personalised oracle feature representations, providing evidence on predicted comfort, cumulative reward, proxy HVAC $\Delta T$, intervention efficiency, action switching and agreement with recorded thermal preferences.
\end{enumerate}
The structure of this paper is as follows. Related work is reviewed and analysed in Section~\ref{sec:related_work}. The proposed personalised thermal comfort architecture is introduced in Section~\ref{sec:framework_overview}. The experimental methodology and results are presented in Section~\ref{sec:results_discussion}. Finally, conclusions and directions for future work are provided in Section~\ref{sec:conclusion}.

\section{Related Work} \label{sec:related_work}This section reviews the literature underpinning the proposed two-stage approach. It first considers conventional and personalised thermal comfort modelling, followed by machine learning approaches that use physiological and environmental sensing to predict individual thermal preferences. It then examines reinforcement learning for intelligent thermal and HVAC decision-making, before identifying the unresolved gap between personalised comfort prediction and sequential intervention-policy generation.

\subsection{Thermal Comfort Modelling}
\label{sec:thermal_comfort_modelling}
Thermal comfort research has historically relied on population‑level models including PMV/PPD~\citep{fanger1970thermal}, which assume steady‑state conditions, fixed metabolic rates and uniform environments. Although widely adopted in building standards, PMV has been repeatedly shown to mispredict comfort in real‑world settings due to its inability to capture behavioural adaptation, transient conditions and inter‑individual variability \citep{humphreys2002validity,d2019fifty}. Adaptive comfort models introduced contextual adaptation based on outdoor temperature, yet still operate at the population level and do not incorporate physiological signals \citep{de1998developing}.

Recent advances have shown that thermal comfort is highly individualised, shaped by physiology, acclimatisation, clothing, activity level and behavioural patterns \citep{kim2018personal2, vculic2021smart, abdallah2016sensing}. Wearable sensing technologies now enable continuous measurement of skin temperature, heart rate, activity and peripheral thermal states, providing richer data streams for personalised comfort inference \citep{liu2019personal, abdelrahman2022personal}. Work published between 2020 and 2025 has expanded this direction, consistently demonstrating that personalised comfort models outperform PMV by margins exceeding 30–50\% in accuracy \citep{park2022prediction, vculic2021smart, haruehansapong2023personal}. These findings highlight a clear shift toward occupant‑centric comfort modelling, in which physiological and behavioural signals form the foundation for accurate, adaptive thermal-preference prediction.

\subsection{Machine Learning for Personalised Thermal Comfort Prediction} \label{sec:ml_personalised_thermal_comfort}
Machine learning has become the dominant approach for personalised thermal comfort prediction, specifically when applied to multimodal physiological and environmental data. Ensemble learning methods \gls*{rf}, \gls*{gb} and \gls*{et} have demonstrated strong predictive performance on wearable datasets, achieving accuracies between 0.65 and 0.85 across diverse occupants \citep{liu2019personal,abdelrahman2022personal}. These models remain competitive in recent studies due to their robustness on tabular sensor data \citep{vculic2021smart, rane2023enhancing}.

Deep learning approaches have also gained traction. Convolutional and recurrent neural networks have been used to capture temporal dynamics in physiological signals \citep{haruehansapong2023personal, sahoh2026deep}, while transformer‑based architectures have recently been explored for modelling long‑range thermal preference dependencies \citep{zhang2023transfer}. Hybrid models that combine physiological sensing with behavioural data \citep{park2022prediction, he2023more} have demonstrated improved generalisation, particularly in residential environments. A notable trend in recent literature (2020–2025) is the shift from discrete thermal sensation labels toward probabilistic comfort representations. Studies including \cite{Gloria} and \cite{sahoh2026deep} show that probability‑based comfort outputs provide richer feedback for downstream control tasks, enabling more stable and interpretable decision‑making. However, most machine learning studies focus exclusively on prediction rather than control, leaving a clear gap in integrating personalised comfort inference with real‑time HVAC decision‑making.

\subsection{Intelligent Thermal Control and Reinforcement Learning}
\label{sec:intelligent_thermal_control}
Intelligent \gls*{hvac} control has been widely explored using optimisation, model predictive control (MPC), and \gls*{rl}. MPC approaches have demonstrated strong performance in energy-efficient building control \citep{afram2014theory, ma2014stochastic, kim2022energy}, but require accurate thermal models and often struggle with occupant variability. RL offers a model‑free alternative capable of learning optimal control policies directly from interaction data. Earlier work has also explored physiologically controlled HVAC systems in which individual thermal preferences and physiological measurements were combined to determine consensus temperature set-points for shared environments. While these approaches demonstrated the feasibility of occupant-centric control, they relied on predefined control algorithms rather than adaptive reinforcement learning policies \citep{ihianle2022towards}.

Recent \gls*{rl}‑\gls*{hvac} studies have applied \gls*{dqn}, Soft Actor–Critic (SAC), Proximal Policy Optimisation (PPO) and multi‑agent \gls*{rl} to building control \citep{yu2021review, vazquez2020multi, gupta2021energy, wei2023deep}. These approaches demonstrate improved energy efficiency and comfort maintenance compared to rule‑based controllers. Multi‑agent \gls*{rl} has been used for multi‑zone buildings \citep{vazquez2020multi}, while deep \gls*{rl} has been applied to dynamic occupancy‑aware control \citep{wei2023deep}. However, nearly all \gls*{rl}-based thermal comfort studies rely on population‑level comfort models, simulated thermal sensation labels or building‑level environmental data. Even recent works from 2020 to 2025 continue to optimise comfort using PMV, thermal sensation votes, or simplified reward functions \citep{yu2021review, gupta2021energy, wei2023deep}. Very few studies integrate personalised comfort predictions, and none embed probabilistic comfort outputs from personalised oracles directly into the \gls*{rl} state representation. Furthermore, most \gls*{rl}-\gls*{hvac} studies assume access to indoor air temperature, HVAC setpoints or building thermal models data unavailable in wearable‑sensor datasets, for example \citep{liu2019personal}. This limits the applicability of existing RL approaches to personalised, physiology‑driven comfort control.

Recent studies have shown a clear trajectory toward personalised, data‑driven thermal comfort modelling, yet several limitations persist. While advances in the application of machine learning have demonstrated the value of personalised comfort prediction using physiological sensing, they remain largely decoupled from real‑time \gls*{hvac} control. Conversely, \gls*{rl}‑based \gls*{hvac} controllers have shown strong potential for adaptive building control, but continue to rely on population‑level comfort models, simulated environments or building‑level data. This disconnect between personalised comfort inference and sequential decision‑making represents a critical gap. Existing \gls*{rl} controllers do not incorporate probabilistic comfort outputs from personalised models, nor do they operate effectively in wearable‑sensor environments where indoor temperature and HVAC setpoints are unavailable. Recent studies between 2020 and 2025 highlight the need for occupant‑centric \gls*{rl} frameworks capable of integrating physiological sensing, personalised comfort prediction and adaptive control within a unified architecture.

The paper addresses these challenges by proposing a two-stage personalised thermal comfort approach that couples occupant-specific Comfort Oracles with reinforcement learning-based intervention-policy generation. By embedding probabilistic comfort predictions directly into the \gls*{rl} state representation and evaluating multiple \gls*{rl} paradigms under identical personalised conditions, the study advances personalised thermal decision-making beyond prediction alone.

\section{The Proposed Two-Stage Architecture} \label{sec:framework_overview} 
\begin{figure*}[htb!]
\centering
\includegraphics[width=\linewidth]{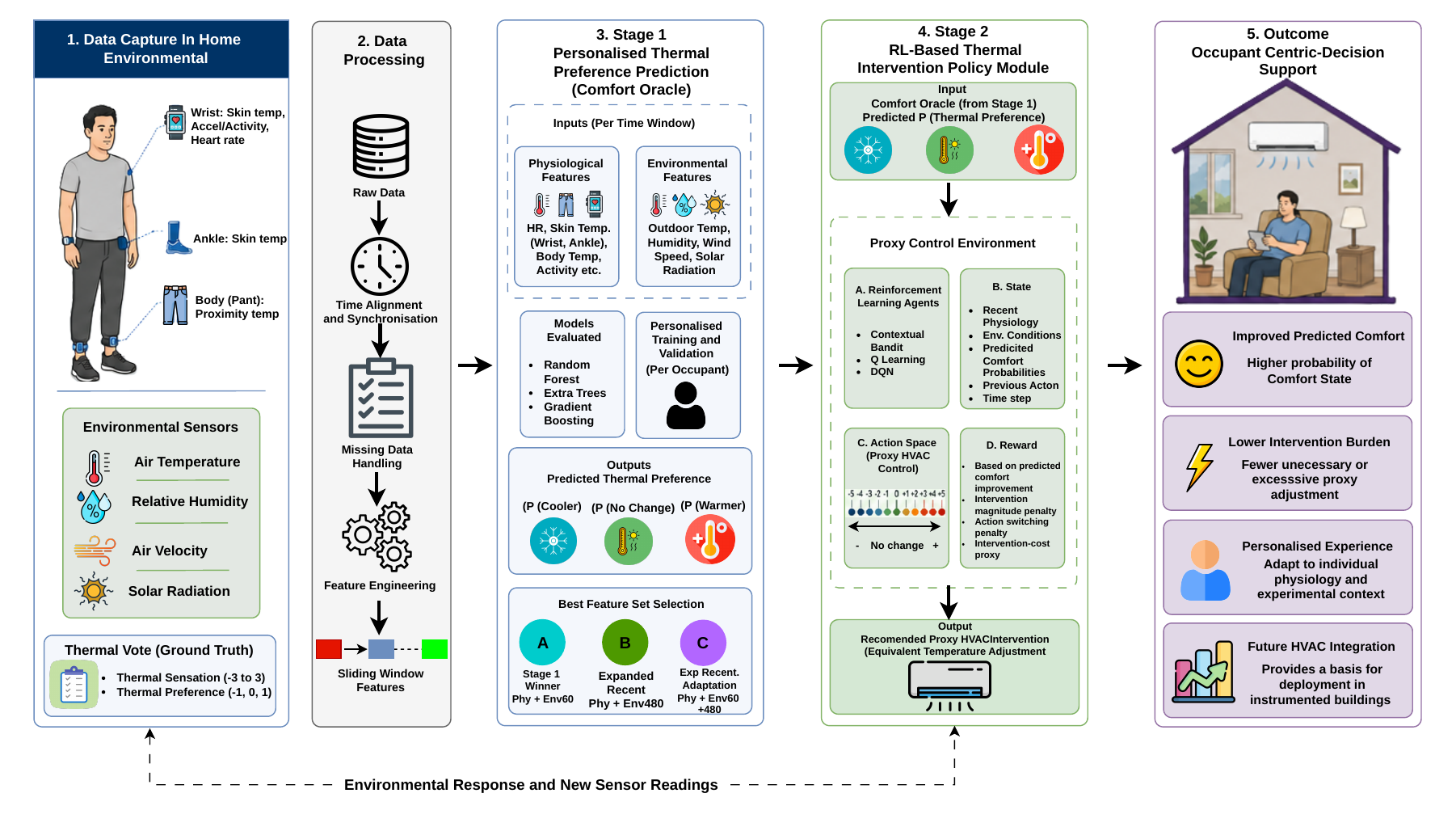}
\caption{Proposed two-stage personalised thermal comfort architecture. Stage 1 develops a personalised comfort oracle for thermal preference prediction using physiological, body-proximity and weather-derived features. Stage 2 embeds the comfort oracle within a reinforcement learning control module to recommend an equivalent environmental-temperature intervention that maximises predicted comfort while penalising unnecessary or excessive control actions.}
\label{fig:proposed_framework}
\end{figure*}

This paper proposes a two-stage personalised thermal comfort architecture that integrates occupant-specific thermal preference prediction with \gls*{rl}-based intervention-policy generation. The proposed architecture, as shown in Figure~\ref{fig:proposed_framework}, integrates personalised machine learning and sequential decision-making to recommend adaptive environmental-temperature interventions for individual occupants. The motivation is that prediction alone is insufficient for intelligent comfort management: a model may estimate whether an occupant prefers a cooler, warmer or unchanged environment, but an additional decision-making mechanism is required to determine the most appropriate control action. The proposed architecture therefore separates the problem into two connected tasks: first, learning an occupant-specific comfort model; and second, using this model to guide adaptive environmental-temperature interventions.

In Stage~1, a personalised thermal preference model, referred to as the \emph{comfort oracle}, is developed for each occupant. The oracle uses time-windowed physiological features, including heart rate, wrist skin temperature, ankle skin temperature and body-proximity temperature, together with environmental sensing including outdoor temperature, humidity, wind speed and solar radiation. These variables follow the structure of previous wearable-sensor-based personal comfort modelling studies, where physiological and environmental signals are used to infer individual thermal preference rather than population-average comfort responses~\citep{liu2019personal,abdelrahman2022personal}. The output of the comfort oracle is not only a discrete thermal preference label, but also a probability distribution over three possible preference states: cooler, no change and warmer. This probabilistic output provides a compact representation of the occupant's current comfort state and becomes a key input to the control stage.

In Stage~2, the comfort oracle is embedded within a reinforcement learning approach. At each decision step, the controller observes the current physiological state, weather-derived context, body-proximity thermal state and oracle-predicted comfort probabilities. It then selects an equivalent environmental-temperature control action from a discrete action space. Since the dataset does not contain directly measured indoor air temperature or HVAC setpoint data, this action is interpreted as a simulated environmental-temperature control proxy rather than a physical thermostat command. This proxy represents the magnitude and direction of local thermal intervention that would be required to move the occupant towards a more comfortable state.

The architecture evaluates three decision-making approaches: a Contextual Bandit, Q-learning and a Deep Q-Network (DQN). The Contextual Bandit provides a strong one-step optimisation baseline, while Q-learning and DQN evaluate whether sequential decision-making can improve control by considering future comfort rewards. By coupling a personalised comfort oracle with reinforcement learning, the proposed architecture extends personal comfort modelling from a predictive task into a prescriptive control approach. This provides a basis for adaptive, occupant-centred thermal management in future intelligent home and workplace environments. Unlike our earlier physiologically controlled HVAC framework, which employed algorithmic consensus-based temperature adjustment, the proposed architecture integrates participant-specific Comfort Oracles with reinforcement learning to enable adaptive intervention-policy generation \citep{ihianle2022towards}.

\section{Experimental Methodology} \label{sec:methodology} This section describes the experimental methodology used to develop and evaluate the proposed two-stage personalised thermal comfort approach. The evaluation follows the architecture presented in Figure~\ref{fig:proposed_framework}. Stage~1 develops participant-specific Comfort Oracles that predict individual thermal preference from physiological and environmental observations, while Stage~2 embeds these personalised oracles within three \gls*{rl} controllers - \gls*{cb}, \gls*{ql} and  \gls*{dqn} to investigate how different decision-making strategies translate personalised comfort predictions into adaptive proxy intervention policies. To ensure a fair comparison, identical participant-specific training, validation and testing partitions are maintained throughout both stages. All experiments are implemented in Python using Scikit-learn for personalised comfort prediction and PyTorch for deep reinforcement learning.

\subsection{Dataset and Feature Engineering} \label{sec:dataset} The proposed approach was evaluated using the publicly available personalised thermal comfort dataset introduced by Liu et. al., \citep{liu2019personal}. The dataset contains wearable physiological measurements with self-reported thermal preference votes collected from participants during everyday activities. The physiological variables include heart rate, wrist skin temperature, ankle skin temperature and body-proximity temperature, while environmental data comprises outdoor air temperature, relative humidity, wind speed and solar radiation obtained from a nearby weather station. Thermal preference is represented using three classes corresponding to a preference for a cooler ($-1$), no change ($0$) or warmer ($+1$) environment.

Following the data preparation procedure by Liu et. al., \cite{liu2019personal}, four pre-constructed feature representations are evaluated: physiological features only (Phys), physiological features combined with 60-minute environmental descriptors (Phys+60), physiological features combined with 480-minute environmental descriptors (Phys+480), and a combined representation integrating both environmental data (Phys+60+480). These feature representations served as candidate inputs to the personalised Comfort Oracles developed in Stage~1 and are subsequently evaluated within the reinforcement learning controllers in Stage~2.

Experiments were performed independently for each participant to preserve the personalised nature of thermal comfort modelling. For each participant, the samples were retained in chronological order and partitioned into training (70\%), validation (10\%), and testing (20\%) subsets, following the protocol of \cite{liu2019personal}. No random shuffling or stratified sampling was applied. The same participant-specific partitions were maintained throughout both stages to ensure consistent evaluation of personalised prediction and adaptive thermal control performance.

\subsection{Stage 1: Personalised Thermal Preference Prediction}
\label{sec:stage1}
Stage~1 develops participant-specific Comfort Oracles to estimate individual thermal preference from the feature representations described in Section~\ref{sec:dataset}. Thermal preference prediction is formulated as a three-class supervised classification problem, where the classes represent a preference for cooler ($-1$), no change ($0$) or warmer ($+1$) conditions. Models are trained independently for each participant to account for inter-individual differences in physiological response, environmental exposure and thermal adaptation.

For participant $p$, let $\mathbf{x}^{(r)}_{p,t}\in\mathbb{R}^{m_r}$ denote the feature vector at time step $t$ under representation:

\begin{equation}
r\in\{\text{Phys},\text{Phys+60},\text{Phys+480},\text{Phys+60+480}\}
\end{equation}
The corresponding Comfort Oracle learns the mapping

\begin{equation}
f^{(r)}_{p}:\mathbf{x}^{(r)}_{p,t}
\rightarrow
\mathbf{p}^{(r)}_{p,t},
\label{eq:comfort_oracle_mapping}
\end{equation}

\noindent where

\begin{equation}
\mathbf{p}^{(r)}_{p,t}
=
\left[
P(y=-1\mid\mathbf{x}^{(r)}_{p,t}),
P(y=0\mid\mathbf{x}^{(r)}_{p,t}),
P(y=+1\mid\mathbf{x}^{(r)}_{p,t})
\right]
\label{eq:comfort_probability}
\end{equation}

\noindent is the predicted probability distribution over the three thermal preference classes. The probabilities sum to one, and the neutral-class probability provides a continuous estimate of the likelihood that no environmental adjustment is required. This probabilistic output forms the principal interface with the Stage~2 controllers.

Three tree-based ensemble models are evaluated as candidate Comfort Oracles: \gls*{rf}, \gls*{gb} and \gls*{et}. These models are selected because they are well suited to heterogeneous tabular sensor data and can represent nonlinear interactions between physiological and environmental variables without strong distributional assumptions~\citep{breiman2001random,geurts2006extremely,friedman2001greedy}. For each participant and feature representation, candidate models are trained on the participant-specific training partition and ranked on the validation partition. To account for class imbalance and avoid selecting an oracle solely on overall accuracy, the validation criterion combined accuracy, balanced accuracy, Macro F$_1$ and Cohen's $\kappa$:

\begin{equation}
S(M_i)
=
\frac{1}{4}
\left(
\mathrm{Acc}_i+
\mathrm{BalAcc}_i+
\mathrm{MacroF1}_i+
\kappa_i
\right),
\label{eq:oracle_score}
\end{equation}

\noindent where $M_i$ denotes a candidate model. The selected Comfort Oracle was

\begin{equation}
M^{*(r)}_{p}
=
\arg\max_{M_i\in\mathcal{M}}
S(M_i),
\label{eq:oracle_selection}
\end{equation}

\noindent with $\mathcal{M}=\{\mathrm{RF},\mathrm{GB},\mathrm{ET}\}$. The selected model was then retrained using the combined training and validation partitions and evaluated once on the held-out test partition.

This procedure produces one selected Comfort Oracle for every participant - feature representation combination. Stage~1 therefore evaluates both model-specific and feature-specific predictive performance, while Stage~2 uses the selected environmental-context oracles (Phys+60, Phys+480 and Phys+60+480) to compare how \gls*{cb}, \gls*{ql} and \gls*{dqn} translate personalised comfort probabilities into adaptive intervention recommendations.

\subsection{Stage 2: Reinforcement Learning-Based Thermal Control}\label{sec:stage2}
Stage~2 extends personalised thermal preference prediction into adaptive decision-making by embedding the participant-specific Comfort Oracles from Stage~1 within three reinforcement learning controllers: \gls*{cb}, \gls*{ql} and \gls*{dqn}. Rather than proposing new \gls*{rl} algorithms, this work introduces a personalised thermal control formulation that integrates participant-specific probabilistic Comfort Oracles with standard \gls*{rl} controllers through a unified state representation and reward design. This enables existing RL methods to operate directly on personalised comfort probabilities rather than discrete thermal preference labels. This extends the prediction-focused approach of \cite{liu2019personal} into a closed-loop, human-in-the-loop proxy control setting, in which participant-specific thermal preferences guide the evaluation of personalised intervention policies. In the context of this study, thermal control refers to the generation and evaluation of equivalent environmental-temperature control signals within a proxy closed-loop environment. The proxy is used as an operational representation of the direction and relative magnitude of the intervention that could subsequently be translated into thermostat or HVAC actuation. This abstraction is necessary because the dataset does not contain measured indoor air temperature, thermostat setpoints, actuator states, building thermal dynamics or \gls*{hvac} power consumption. Accordingly, the Stage~2 experiments evaluate personalised control-policy behaviour under a common proxy environment rather than direct actuation of a physical HVAC system.

To investigate how different sequential decision-making paradigms influence personalised thermal control, three representative reinforcement learning approaches are evaluated. \gls*{cb} optimise immediate reward without considering future consequences, providing a strong one-step decision baseline. \gls*{ql} extends this by learning long-term state-action values through temporal-difference learning, while \gls*{dqn} replaces the tabular value function with a neural approximation capable of modelling more complex state representations \citep{watkins1992q,mnih2015human}. These controllers integrate increasing levels of sequential reasoning and function approximation while operating under the same personalised state and reward formulation.

For participant $p$ and feature representation $r$, the state observed at decision step $t$ is, 

\begin{equation}
s^{(r)}_{p,t}
=
\left[
\mathbf{x}^{(r)}_{p,t},
\mathbf{p}^{(r)}_{p,t},
a_{p,t-1}
\right],
\label{eq:rl_state}
\end{equation}

\noindent where $\mathbf{x}^{(r)}_{p,t}$ is the current physiological and environmental feature vector, $\mathbf{p}^{(r)}_{p,t}$ is the personalised Comfort Oracle probability vector defined in Equation~\ref{eq:comfort_probability}, and $a_{p,t-1}$ is the previous control action. The inclusion of the previous action provides short-term control context and allows rapid changes between successive interventions to be penalised.

The controller selects an action from

\begin{equation}
\mathcal{A}
=
\{-5,-4,-3,-2,-1,0,+1,+2,+3,+4,+5\},
\label{eq:rl_action_space}
\end{equation}

\noindent where negative, zero and positive values represent a decrease, no change or increase in the equivalent environmental temperature, respectively. These values operationalise the proxy control signal defined above and should be interpreted as relative intervention levels rather than direct thermostat setpoints or realised temperature changes. Following action selection, the proxy thermal state is updated and re-evaluated by the participant-specific Comfort Oracle. The resulting reward balances predicted comfort against intervention magnitude, switching behaviour and the associated energy proxy:

\begin{equation}
R_{p,t}
=
w_c p^{\mathrm{neutral}}_{p,t+1}
-
w_a |a_{p,t}|
-
w_s \mathbb{I}(a_{p,t}\neq a_{p,t-1})
-
w_e E_{p,t},
\label{eq:rl_reward}
\end{equation}

\noindent where $p^{\mathrm{neutral}}_{p,t+1}$ is the post-intervention probability that no further temperature change is required, $|a_{p,t}|$ represents intervention magnitude, $\mathbb{I}(\cdot)$ penalises action switching, and $E_{p,t}$ is an intervention-based energy proxy. The same state, action and reward definitions are used across controller families to ensure that differences in performance are attributable to the decision-making strategy rather than to different control objectives. These reward components are reflected in the Stage 2 evaluation metrics (Section~\ref{sec:evaluation_metrics}), enabling controller behaviour to be analysed in terms of comfort, intervention magnitude, efficiency and behavioural stability.

\paragraph{\textbf{Contextual Bandit:}}The \gls*{cb} controller provides a myopic decision-making baseline. At each step, it selects the action with the highest estimated immediate reward under the current personalised state,

\begin{equation}
a^{*}_{p,t}
=
\arg\max_{a\in\mathcal{A}}
\mathbb{E}
\left[
R_{p,t}\mid s_{p,t},a
\right].
\label{eq:cb_policy}
\end{equation}

Unlike sequential RL methods, \gls*{cb} does not explicitly model the long-term consequences of an action. It is therefore suited to responsive comfort correction when immediate predicted comfort is prioritised. Three reward-weighting variants are evaluated: Comfort, Balanced and Energy.

\paragraph{\textbf{Q-learning:}}\gls*{ql} is a model-free temporal-difference method that learns the expected long-term value of each state-action pair~\citep{watkins1992q,clifton2020q}. Its action-value estimate is updated as

\begin{equation}
%\begin{aligned}
\begin{gathered}
Q(s_{p,t},a_{p,t}) \leftarrow Q(s_{p,t},a_{p,t})
+ \\
\alpha\left[
R_{p,t}
+\gamma\max_{a\in\mathcal{A}}Q(s_{p,t+1},a)
-Q(s_{p,t},a_{p,t})
\right]
%\end{aligned}
\end{gathered}
\label{eq:q_learning_update}
\end{equation}

\noindent where $\alpha$ is the learning rate and $\gamma$ controls the contribution of future rewards. QL is evaluated using $\alpha\in\{0.05, 0.08, 0.12, 0.16\}$.

\paragraph{\textbf{Deep Q-Network:}}\gls*{dqn} extends QL by approximating the action-value function using a neural network, allowing nonlinear relationships within the personalised physiological, environmental and comfort-probability state to be represented~\citep{mnih2015human}. The network parameters $\theta$ are updated by minimising the temporal-difference loss

\begin{equation}
%\begin{aligned}
\begin{gathered}
L(\theta)
= \\
\mathbb{E}
\left[
\left(
R_{p,t}
+
\gamma\max_{a'}Q(s_{p,t+1},a';\theta^{-}) 
-
Q(s_{p,t},a_{p,t};\theta)
\right)^2
\right],
\label{eq:dqn_loss}
\end{gathered}
%\end{aligned}
\end{equation}

\noindent where $\theta^{-}$ denotes the target network parameters. Experience replay and target network updates are employed to improve training stability. The DQN model was evaluated using learning rates $\eta \in \{0.0001, 0.0002, 0.0003, 0.0005\}$.
For each participant, oracle feature representation and controller family, the candidate configurations are trained using the participant-specific training data and ranked using mean validation reward. The highest-ranked variant is then evaluated on the held-out test partition. This produced directly comparable \gls*{cb}, \gls*{ql} and \gls*{dqn} policies for Phys+60, Phys+480 and Phys+60+480, as reported in Section~\ref{sec:stage2_results}. This experimental design isolates the influence of the reinforcement learning policy while keeping the underlying personalised comfort representation fixed, enabling a fair comparison of CB, QL and DQN under identical participant-specific conditions.

Algorithm~\ref{alg:personalised_rl_control} summarises the proposed personalised thermal control procedure, illustrating how the participant-specific Comfort Oracles developed in Stage~1 interact with the RL controller to generate adaptive equivalent environmental-temperature interventions.

\begin{algorithm}[t]
\caption{Personalised Comfort-Oracle-Guided Thermal Control}
\label{alg:personalised_rl_control}
\small

\KwIn{
Participant-specific Comfort Oracle $M^{*(r)}_{p}$;
participant observations $\mathbf{x}^{(r)}_{p,t}$;
controller family $C\in\{\mathrm{CB},\mathrm{QL},\mathrm{DQN}\}$;
action space $\mathcal{A}$
}

\KwOut{Personalised control policy $\pi^{*(r)}_{p,C}$}

Initialise controller $C$ and previous action $a_{p,0}=0$\;

\For{each decision step $t$}
{
Obtain the Comfort Oracle probabilities
$\mathbf{p}^{(r)}_{p,t}
=
M^{*(r)}_{p}(\mathbf{x}^{(r)}_{p,t})$\;

Construct the personalised control state
$s^{(r)}_{p,t}
=
[\mathbf{x}^{(r)}_{p,t},
\mathbf{p}^{(r)}_{p,t},
a_{p,t-1}]$\;

Select $a_{p,t}\in\mathcal{A}$ using the current controller policy\;

Apply $a_{p,t}$ to the simulated environmental-temperature proxy\;

Obtain the updated observation
$\mathbf{x}^{(r)}_{p,t+1}$ and re-evaluate it using
$M^{*(r)}_{p}$\;

Compute $R_{p,t}$ using Equation~\ref{eq:rl_reward}\;

Update the controller according to the CB, QL or DQN learning rule\;
}

Select the highest-reward validated controller configuration\;

Evaluate the selected policy on the held-out test partition\;

\Return{$\pi^{*(r)}_{p,C}$}
\end{algorithm}

\subsection{Evaluation Metrics} \label{sec:evaluation_metrics}
Since the proposed approach comprises two distinct stages, different evaluation criteria are required. Stage~1 is formulated as a supervised multiclass classification problem and is therefore evaluated using predictive performance metrics. Stage~2 is formulated as a sequential decision-making problem in which the objective is to maximise personalised comfort while maintaining efficient and stable proxy control. The evaluation is formulated to focus on controller behaviour.

Stage~1 performance was assessed using Accuracy, Balanced Accuracy, Macro F$_1$ and Cohen's $\kappa$. Accuracy measures overall predictive correctness, while Balanced Accuracy and Macro F$_1$ provide class-balanced evaluation under the unequal thermal preference distributions observed across participants. Cohen's $\kappa$ further measures agreement beyond chance.

Stage~2 performance is evaluated using mean comfort probability, mean reward, proxy HVAC temperature adjustment (HVAC $\Delta T$), reward efficiency, comfort efficiency and action switching rate. Mean comfort probability quantifies the controller's ability to maintain the occupant in the predicted ``No Change'' state. HVAC $\Delta T$ denotes the mean magnitude of the equivalent environmental-temperature control signal recommended by the controller. It therefore represents a proxy for the direction and relative magnitude of prospective HVAC intervention, rather than a measured thermostat setpoint change, realised indoor-temperature change or physical actuator response. HVAC $\Delta T$ and action switching rate characterise intervention magnitude and policy stability, respectively. Reward efficiency and comfort efficiency measure the reward and predicted comfort achieved per unit of proxy HVAC intervention. These efficiency measures are therefore relative policy-comparison metrics and should not be interpreted as measured building energy efficiency or absolute energy savings.

Action agreement with the recorded thermal preference labels is additionally assessed using Accuracy, Balanced Accuracy, Macro F$_1$ and Cohen's $\kappa$, providing an interpretable measure of how closely controller actions aligned with the occupant's observed preference.

Non-parametric Friedman tests are used to compare the controllers across participants for each Stage~2 performance metric and oracle feature representation. Where the omnibus test is significant, pairwise Wilcoxon signed-rank tests with Holm correction are applied to identify the controller pairs responsible for the differences.

\section{Experiments and Results} \label{sec:results_discussion}
To evaluate the proposed personalised thermal comfort approach, the experiments followed the sequential approach as depicted in Figure~\ref{fig:proposed_framework}. Stage~1 assesses the predictive performance of personalised Comfort Oracles using multiple ensemble learning models and feature representations. Stage~2 embeds the selected Comfort Oracles within the \gls*{rl} module, where \gls*{cb}, \gls*{ql} and \gls*{dqn} controllers are evaluated on unseen participant data. Prediction performance is evaluated using accuracy, Macro F$_1$ and Cohen's $\kappa$, while control performance is assessed using comfort probability, reward, intervention magnitude, reward efficiency and action agreement.

\subsection{Stage 1 Personalised Thermal Preference Prediction}\label{sec:stage1_results}
Stage~1 identifies the most appropriate Comfort Oracle for each participant by evaluating three ensemble learning models - \gls*{rf}, \gls*{gb} and \gls*{et}. Each model was trained and evaluated using four feature representations: physiological features only (Phys), physiological features with 60-minute environmental descriptors (Phys+60), physiological features with 480-minute environmental descriptors (Phys+480), and the combined environmental representation (Phys+60+480). Candidate models are evaluated, with the highest-ranked model selected as the participant-specific Comfort Oracle for Stage~2 thermal control. This participant-specific selection strategy acknowledges the substantial inter-individual variability in thermal preference and is consistent with recent personalised thermal comfort studies \citep{liu2019personal,kim2018personal,kim2018personal2}.

\subsubsection{Candidate Comfort Oracle Performance} 
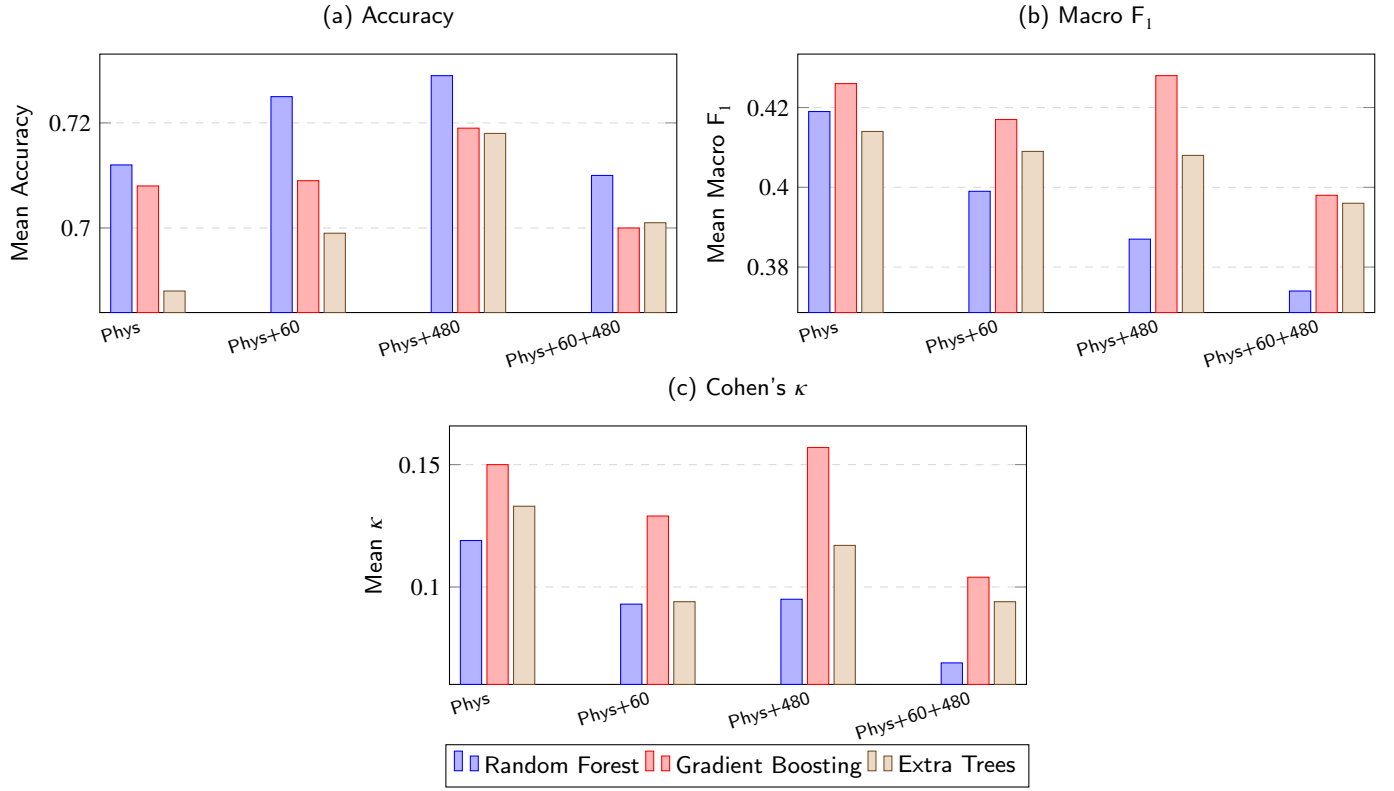
\begin{figure*}[t]
\centering
\begin{tikzpicture}
\begin{groupplot}[
group style={group size=2 by 2,
horizontal sep=1.6cm,
vertical sep=1.5cm},
width=0.53\textwidth,
height=5.0cm,
ybar,
/pgf/bar width=8pt,
major x tick style=transparent,
ymajorgrids=true,
grid style={dashed,gray!30},
scaled y ticks=false,
xtick=data,
symbolic x coords={Phys,Phys60,Phys480,Phys60480},
xticklabels={Phys,Phys+60,Phys+480,Phys+60+480},
x tick label style={font=\scriptsize,rotate=18,anchor=east},
legend style={
at={(1.105,-1.670)},
anchor=north,
legend columns=3,
draw=black,
fill=white,
font=\small
}
]
\nextgroupplot[title={(a) Accuracy},ylabel={Mean Accuracy}]
\addplot coordinates {(Phys,0.712) (Phys60,0.725) (Phys480,0.729) (Phys60480,0.710)};
\addlegendentry{Random Forest}
\addplot coordinates {(Phys,0.708) (Phys60,0.709) (Phys480,0.719) (Phys60480,0.700)};
\addlegendentry{Gradient Boosting}
\addplot coordinates {(Phys,0.688) (Phys60,0.699) (Phys480,0.718) (Phys60480,0.701)};
\addlegendentry{Extra Trees}
\nextgroupplot[title={(b) Macro F$_1$},ylabel={Mean Macro F$_1$}]
\addplot coordinates {(Phys,0.419) (Phys60,0.399) (Phys480,0.387) (Phys60480,0.374)};
\addplot coordinates {(Phys,0.426) (Phys60,0.417) (Phys480,0.428) (Phys60480,0.398)};
\addplot coordinates {(Phys,0.414) (Phys60,0.409) (Phys480,0.408) (Phys60480,0.396)};
\nextgroupplot[hide axis]
\nextgroupplot[title={(c) Cohen's $\kappa$},ylabel={Mean $\kappa$},xshift=-0.265\textwidth]
\addplot coordinates {(Phys,0.119) (Phys60,0.093) (Phys480,0.095) (Phys60480,0.069)};
\addplot coordinates {(Phys,0.150) (Phys60,0.129) (Phys480,0.157) (Phys60480,0.104)};
\addplot coordinates {(Phys,0.133) (Phys60,0.094) (Phys480,0.117) (Phys60480,0.094)};
\end{groupplot}
\end{tikzpicture}
\caption{Comparison of the candidate Stage~1 Comfort Oracle models across the four feature representations. Mean prediction performance is reported using (a) Accuracy, (b) Macro F$_1$-score and (c) Cohen's $\kappa$, averaged over all participants. Phys denotes physiological features only; Phys+60 denotes physiological features with 60-minute environmental descriptors; Phys+480 denotes physiological features with 480-minute environmental descriptors; and Phys+60+480 combines both environmental contexts.}
\label{fig:stage1_performance}
\end{figure*}

Figure~\ref{fig:stage1_performance} compares the predictive performance of the three candidate Comfort Oracle models across the four feature representations. Overall, the three ensemble classifiers achieved comparable performance, with mean accuracies ranging from approximately 0.69 to 0.73. Integrating environmental features provided only modest improvements over physiological features alone, with the Phys+480 representation consistently achieving the highest mean accuracy for all three classifiers. The relatively small performance differences suggest that physiological signals capture much of the information required to predict individual thermal preferences, while longer-term environmental context provides complementary rather than dominant predictive information.

A more informative comparison is provided by Macro F$_1$ and Cohen's $\kappa$, which are less sensitive to the class imbalance inherent in personalised thermal preference data. \gls*{gb} consistently achieved the highest Macro F$_1$ and Cohen's $\kappa$, indicating better discrimination across the three thermal preference classes, whereas \gls*{rf} generally produced the highest overall accuracy. \gls*{et} remained competitive across all feature representations but exhibited greater variability. The performance characteristics indicate that no single classifier consistently dominates across all participants or evaluation metrics, reinforcing the use of participant-specific model selection rather than a single global prediction model.

While the aggregated results provide an overall comparison of the candidate models, they conceal substantial inter-participant variability. Table~\ref{tab:stage1_full_physiology_env480_acc_f1_kappa} illustrates the participant-level performance for the strongest-performing feature representation (Phys+480), with the complete results for all feature representations provided in Appendix~A. 

\begin{table*}[t]
\centering
\scriptsize
\setlength{\tabcolsep}{3pt}
\renewcommand{\arraystretch}{1.0}
\caption{Participant-level Stage~1 performance for Physiology + Env480. Values are Accuracy/Macro F$_1$/Cohen's $\kappa$.}
\label{tab:stage1_full_physiology_env480_acc_f1_kappa}
\begin{tabular}{lcccccccccccccc}
\toprule
	\textbf{Model} & \textbf{1} & \textbf{2} & \textbf{3} & \textbf{4} & \textbf{5} & \textbf{6} & \textbf{7} & \textbf{8} & \textbf{9} & \textbf{10} & \textbf{11} & \textbf{12} & \textbf{13} & \textbf{14} \\
\midrule
RF & \makecell{0.61/\\0.39/\\0.20} & \makecell{0.61/\\0.32/\\0.04} & \makecell{0.65/\\0.39/\\0.15} & \makecell{0.84/\\0.46/\\-0.09} & \makecell{0.84/\\0.42/\\0.27} & \makecell{0.59/\\0.35/\\0.15} & \makecell{0.75/\\0.34/\\0.11} & \makecell{0.89/\\0.31/\\0.00} & \makecell{0.82/\\0.41/\\0.23} & \makecell{0.69/\\0.34/\\0.07} & \makecell{0.88/\\0.31/\\-0.03} & \makecell{0.53/\\0.51/\\0.08} & \makecell{0.68/\\0.55/\\0.14} & \makecell{0.83/\\0.30/\\0.00} \\
\midrule
GB & \makecell{0.68/\\0.43/\\0.32} & \makecell{0.59/\\0.54/\\0.18} & \makecell{0.69/\\0.49/\\0.29} & \makecell{0.91/\\0.32/\\-0.01} & \makecell{0.82/\\0.42/\\0.23} & \makecell{0.39/\\0.27/\\0.06} & \makecell{0.70/\\0.37/\\0.11} & \makecell{0.89/\\0.40/\\0.17} & \makecell{0.76/\\0.35/\\0.05} & \makecell{0.67/\\0.43/\\0.16} & \makecell{0.86/\\0.42/\\0.10} & \makecell{0.62/\\0.61/\\0.25} & \makecell{0.66/\\0.59/\\0.18} & \makecell{0.83/\\0.36/\\0.11} \\
\midrule
ET & \makecell{0.65/\\0.43/\\0.30} & \makecell{0.53/\\0.41/\\0.07} & \makecell{0.65/\\0.42/\\0.18} & \makecell{0.78/\\0.44/\\-0.11} & \makecell{0.85/\\0.47/\\0.38} & \makecell{0.53/\\0.38/\\0.21} & \makecell{0.70/\\0.34/\\0.06} & \makecell{0.89/\\0.31/\\0.00} & \makecell{0.78/\\0.35/\\0.06} & \makecell{0.67/\\0.39/\\0.11} & \makecell{0.86/\\0.31/\\-0.04} & \makecell{0.65/\\0.64/\\0.31} & \makecell{0.68/\\0.52/\\0.10} & \makecell{0.83/\\0.30/\\0.00} \\
\bottomrule
\end{tabular}
\end{table*}

Considerable variations are evident across participants, with accuracies ranging from approximately 0.39 to 0.93 and corresponding differences in Macro F$_1$ and Cohen's $\kappa$. For example, Participant~4 achieved consistently high accuracies across all classifiers, whereas Participants~6 and~12 exhibited substantially lower and more variable performance, reflecting differences in individual thermal preference patterns and class distributions. These variabilities demonstrate how participants respond to thermal changes, reinforcing the need for personalised Comfort Oracles rather than a population-wide predictor.

\subsubsection{Selected Personalised Comfort Oracles}
Table~\ref{tab:stage1_selected_oracle_by_subject} summarises the personalised Comfort Oracle selected for each participant under the four feature representations using the composite validation score defined in Equation~(\ref{eq:oracle_score}). Consistent with the participant-level analysis in Section~\ref{sec:stage1_results}, the selected models vary across both participants and feature representations, with no single learning algorithm consistently providing the highest predictive performance. \gls*{rf} was selected most frequently for participants exhibiting more stable thermal preference patterns, whereas \gls*{gb} and \gls*{et} are selected for participants with greater class imbalance or more heterogeneous preference distributions. This shows that thermal preference prediction is inherently participant-dependent and reinforces the need for personalised Comfort Oracles rather than a single population-wide predictor, consistent with the personalised comfort modelling paradigm reported by \cite{kim2018personal2,liu2019personal,kim2018personal}.

\begin{table*}
\centering
\scriptsize
\caption{Selected personalised Comfort Oracle for each participant and feature representation. Values are Model: Accuracy/Macro F$_1$/Cohen's $\kappa$.}
\label{tab:stage1_selected_oracle_by_subject}
\begin{tabular}{lllll}
\toprule
FeatureShort & Phys & Phys+60 & Phys+480 & Phys+60+480 \\
ID &  &  &  &  \\
\midrule
1 & GB: 0.68/0.45/0.36 & ET: 0.58/0.39/0.17 & ET: 0.65/0.43/0.30 & ET: 0.55/0.36/0.08 \\
2 & ET: 0.39/0.36/-0.05 & ET: 0.49/0.37/-0.00 & ET: 0.53/0.41/0.07 & ET: 0.57/0.41/0.05 \\
3 & ET: 0.62/0.38/0.11 & GB: 0.69/0.51/0.31 & GB: 0.69/0.49/0.29 & GB: 0.69/0.49/0.29 \\
4 & RF: 0.93/0.48/0.00 & RF: 0.87/0.47/-0.07 & RF: 0.84/0.46/-0.09 & RF: 0.82/0.45/-0.10 \\
5 & GB: 0.85/0.47/0.38 & GB: 0.82/0.42/0.23 & GB: 0.82/0.42/0.23 & GB: 0.80/0.42/0.20 \\
6 & RF: 0.45/0.36/0.09 & GB: 0.39/0.27/0.07 & GB: 0.39/0.27/0.06 & GB: 0.41/0.29/0.10 \\
7 & ET: 0.61/0.44/0.06 & RF: 0.77/0.38/0.20 & RF: 0.75/0.34/0.11 & RF: 0.76/0.37/0.17 \\
8 & RF: 0.89/0.31/0.00 & RF: 0.89/0.31/0.00 & RF: 0.89/0.31/0.00 & RF: 0.89/0.31/0.00 \\
9 & RF: 0.80/0.36/0.09 & RF: 0.80/0.36/0.09 & ET: 0.78/0.35/0.06 & RF: 0.82/0.41/0.23 \\
10 & RF: 0.65/0.38/0.08 & ET: 0.63/0.33/-0.03 & RF: 0.69/0.34/0.07 & ET: 0.62/0.34/-0.00 \\
11 & ET: 0.79/0.42/0.11 & GB: 0.83/0.30/-0.06 & GB: 0.86/0.42/0.10 & GB: 0.85/0.39/0.09 \\
12 & ET: 0.62/0.42/0.27 & GB: 0.62/0.61/0.25 & GB: 0.62/0.61/0.25 & GB: 0.65/0.64/0.31 \\
13 & ET: 0.73/0.65/0.31 & RF: 0.68/0.55/0.14 & RF: 0.68/0.55/0.14 & ET: 0.66/0.46/0.01 \\
14 & GB: 0.83/0.36/0.11 & RF: 0.83/0.30/0.00 & GB: 0.83/0.36/0.11 & RF: 0.83/0.30/0.00 \\
\bottomrule
\end{tabular}
\end{table*}

Figure~\ref{fig:predictive_power} summarises the predictive performance of the selected Comfort Oracles across the four feature representations. Although Phys+480 achieved the highest mean prediction accuracy (0.715), the corresponding improvements in Macro F$_1$ and Cohen's $\kappa$ are comparatively modest and remained within one standard deviation of the remaining feature representations. Similarly, combining both environmental contexts (Phys+60+480) did not provide a consistent performance advantage over the individual environmental representations. These results suggest that extending the environmental observation window beyond physiological measurements provides only incremental gains in supervised thermal preference prediction, indicating that much of the predictive information is already captured by the physiological features.

The statistical comparison in Table~\ref{tab:stage1_stat_tests} supports these observations. Friedman tests identified no statistically significant differences between the four feature representations for Accuracy ($p=0.969$), Macro F$_1$ ($p=0.180$) or Cohen's $\kappa$ ($p=0.430$), indicating that the choice of environmental representation had only a limited influence on predictive performance. Likewise, no statistically significant differences are observed between the three ensemble learning algorithms across the evaluated metrics, despite the modest performance differences shown in Figure~\ref{fig:stage1_performance}. These findings suggest that the principal source of variation arises from individual thermal preference characteristics rather than from the learning algorithm or feature representation itself.

%\clearpage

\begin{table}[t]
\centering
\scriptsize
\caption{Statistical comparison of Stage~1 prediction performance using Friedman tests across participants.}
\label{tab:stage1_stat_tests}
\begin{tabular}{llccc}
\toprule
Comparison & Metric & Test & Statistic & $p$-value \\
\midrule
Feature sets & Accuracy & Friedman & 0.250 & 0.969 \\
Feature sets & Macro F$_1$ & Friedman & 4.887 & 0.180 \\
Feature sets & Cohen's $\kappa$ & Friedman & 2.758 & 0.430 \\
Models & Accuracy & Friedman & 1.815 & 0.404 \\
Models & Macro F$_1$ & Friedman & 2.286 & 0.319 \\
Models & Cohen's $\kappa$ & Friedman & 3.857 & 0.145 \\
\bottomrule
\end{tabular}
\end{table}

Unlike Liu \textit{et al.}~\cite{liu2019personal}, who selected a single prediction model before subsequent optimisation, the proposed approach preserves all three environmental representations (Phys+60, Phys+480 and Phys+60+480) for Stage~2. This enables the reinforcement learning controllers to be evaluated using participant-specific Comfort Oracles derived from alternative environmental contexts, allowing the downstream influence of feature representation on adaptive thermal control to be investigated independently of supervised prediction performance.

\begin{tikzpicture}
\begin{axis}[
width=\linewidth,
height=5cm,
ybar,
bar width=8pt,
enlarge x limits=0.18,
ylabel={Predictive power},
xlabel={Feature representation},
ymin=0,
ymax=1,
symbolic x coords={Phys,Phys60,Phys480,PhysBoth},
xtick=data,
xticklabels={Phys,Phys+60,Phys+480,Phys+60+480},
xticklabel style={
    font=\scriptsize, rotate=18,
    anchor=north east
},
ylabel style={font=\small},
xlabel style={font=\small},
legend style={
    at={(0.5,-0.29)},
    anchor=north,
    legend columns=3,
    draw=black,
    fill=white,
    font=\scriptsize
},
ymajorgrids=true,
grid style={dashed,gray!30},
error bars/y dir=both,
error bars/y explicit
]

% Cohen's kappa
\addplot+ coordinates{
(Phys,0.1366) +- (0,0.1364)
(Phys60,0.0936) +- (0,0.1258)
(Phys480,0.1224) +- (0,0.1115)
(PhysBoth,0.1027) +- (0,0.1217)
};
\addlegendentry{Mean Cohen's $\kappa$}

\addplot+ coordinates{
(Phys,0.7028) +- (0,0.1594)
(Phys60,0.7070) +- (0,0.1505)
(Phys480,0.7154) +- (0,0.1399)
(PhysBoth,0.7078) +- (0,0.1399)
};
\addlegendentry{Mean Accuracy}
\addplot+ coordinates{
(Phys,0.4168) +- (0,0.0834)
(Phys60,0.3978) +- (0,0.1021)
(Phys480,0.4120) +- (0,0.0948)
(PhysBoth,0.4030) +- (0,0.0917)
};
\addlegendentry{Mean Macro F$_1$}
\end{axis}
\end{tikzpicture}
\captionof{figure}{Predictive power of the selected personalised Comfort Oracles across the four feature representations. Bars represent the mean performance across the fourteen participants, while error bars denote one standard deviation.}
\label{fig:predictive_power}

\subsection{Stage 2 Thermal Control Performance}\label{sec:stage2_results}
Stage~2 evaluates how effectively the selected personalised Comfort Oracles support adaptive thermal control within the proposed reinforcement learning approach. Using the participant‑specific Comfort Oracles derived in Stage~1, three \gls*{rf} controllers - \gls*{cb}, \gls*{ql} and \gls*{dqn} are evaluated under identical experimental conditions using unseen test data. Unlike Stage~1, which focused on prediction accuracy, Stage~2 assesses the ability of each controller to translate personalised thermal preference predictions into effective control actions by balancing occupant comfort, cumulative reward and intervention efficiency. This enables the influence of both the reinforcement learning strategy and the underlying oracle feature representation on closed-loop thermal regulation to be investigated.

\subsubsection{Controller Performance Using the Selected Comfort Oracles}
\label{sec:stage2_controller_performance}
To ensure a fair comparison between reinforcement learning strategies, the highest-performing variant within each controller family is selected using mean cumulative reward. Table~\ref{tab:stage2_controller_performance} summarises the resulting nine controller configurations, while the complete evaluation of all controller variants is provided in Table~\ref{tab:stage2_all_controller_variants}. The selected controllers are compared using occupant comfort probability, cumulative reward, average HVAC intervention magnitude ($\Delta T$), reward efficiency, comfort efficiency and the distribution of control actions.

\begin{table*}
\centering
\scriptsize
\caption{Best Stage~2 controller variants across oracle feature representations. The best variant within each controller family was selected using mean reward.}
\label{tab:stage2_controller_performance}
\begin{tabular}{lllcccccccc}
\toprule
Oracle & Controller & Best Variant & Comfort & Reward & HVAC $\Delta T$  & Reward/Energy & Comfort/Energy & Decrease & No Change & Increase \\
\midrule
Phys+60 & CB & Bandit Comfort & 0.700 & 89.942 & 3.342  & 26.917 & 0.210 & 0.093 & 0.817 & 0.090 \\
Phys+60 & QL& QL alpha=0.16 & 0.668 & 84.442 & 2.628 & 32.132 & 0.254 & 0.019 & 0.978 & 0.003 \\
Phys+60 & DQN & DQN lr=0.0002 & 0.678 & 84.382 & 3.343  & 25.238 & 0.203 & 0.417 & 0.386 & 0.198 \\
Phys+480 & CB & Bandit Comfort & 0.684 & 87.417 & 3.602 & 24.269 & 0.190 & 0.084 & 0.781 & 0.134 \\
Phys+480 & QL & QL alpha=0.16 & 0.658 & 82.764 & 2.642  & 31.324 & 0.249 & 0.019 & 0.978 & 0.003 \\
Phys+480 & DQN & DQN lr=0.0001 & 0.664 & 82.374 & 3.225 & 25.542 & 0.206 & 0.477 & 0.427 & 0.096 \\
Phys+60+480 & CB & Bandit Comfort & 0.696 & 87.382 & 3.548 & 24.627 & 0.196 & 0.103 & 0.795 & 0.102 \\
Phys+60+480 & QL & QL alpha=0.16 & 0.668 & 82.764 & 2.639 & 31.359 & 0.253 & 0.019 & 0.978 & 0.003 \\
Phys+60+480 & DQN & DQN lr=0.0001 & 0.677 & 82.809 & 3.568 & 23.206 & 0.190 & 0.541 & 0.329 & 0.130 \\
\bottomrule
\end{tabular}
\end{table*}

Across all three oracle feature representations, \gls*{cb} consistently achieved the highest comfort probability (0.684 - 0.700) and cumulative reward (87.4 - 89.9), demonstrating its effectiveness in maximising immediate occupant comfort through contextual action selection. However, this performance was at the expense of larger HVAC interventions, with average temperature adjustments exceeding $3.3^{\circ}$C, resulting in lower reward and comfort efficiency than \gls*{ql}. In contrast, \gls*{ql} adopted a markedly more conservative control policy, producing the smallest HVAC interventions (approximately $2.6^{\circ}$C) while achieving the highest reward efficiency (31.3 - 32.1) and comfort efficiency (0.249 - 0.254). This behaviour is reflected in its action distribution, where approximately 98\% of decisions maintained the current temperature, indicating a stable policy that only intervenes when necessary. \gls*{dqn} exhibited an intermediate control strategy. Although the comfort probability remained comparable to \gls*{ql}, it consistently generated larger HVAC interventions and more frequent temperature adjustments than both alternative controllers. With this, \gls*{dqn} achieved lower reward and comfort efficiency than \gls*{ql} without matching the comfort performance of \gls*{cb}. The broader action distribution further indicates a more exploratory control policy, which is expected from value-function approximation during policy learning but results in less conservative temperature regulation. As observed, the influence of the oracle feature representation was comparatively small. The Phys+60, Phys+480 and Phys+60+480 Comfort Oracles produced similar controller performance across all evaluation metrics, indicating that the \gls*{rl} policy is considerably more influential than the specific environmental representation selected during Stage~1. This observation is consistent with the Stage~1 statistical analysis, where only modest differences are observed between feature representations, and suggests that personalised prediction provides a sufficiently robust state representation for downstream reinforcement learning irrespective of the environmental observation window.

\subsubsection{Comfort - Energy Trade-off Analysis} \label{sec:stage2_tradeoff}
\begin{figure*}[t]
    \centering
    \begin{subfigure}[t]{0.49\textwidth}
        \centering
        \includegraphics[width=\textwidth]{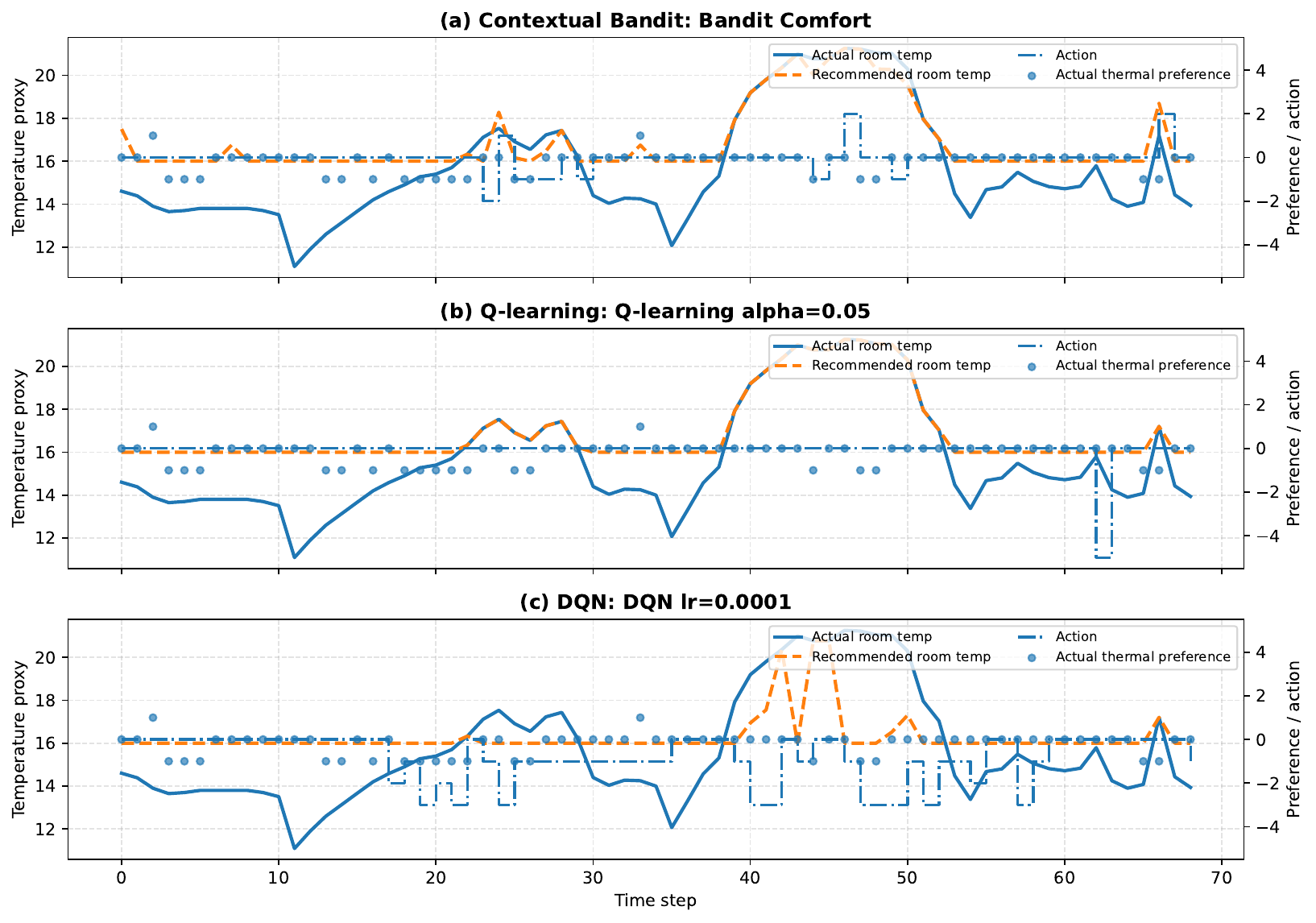}
        \caption{Subject 7: Representative closed-loop thermal regulation using Phys+480}
        \label{fig:subject9}
    \end{subfigure}
    \hfill
    \begin{subfigure}[t]{0.49\textwidth}
        \centering
        \includegraphics[width=\textwidth]{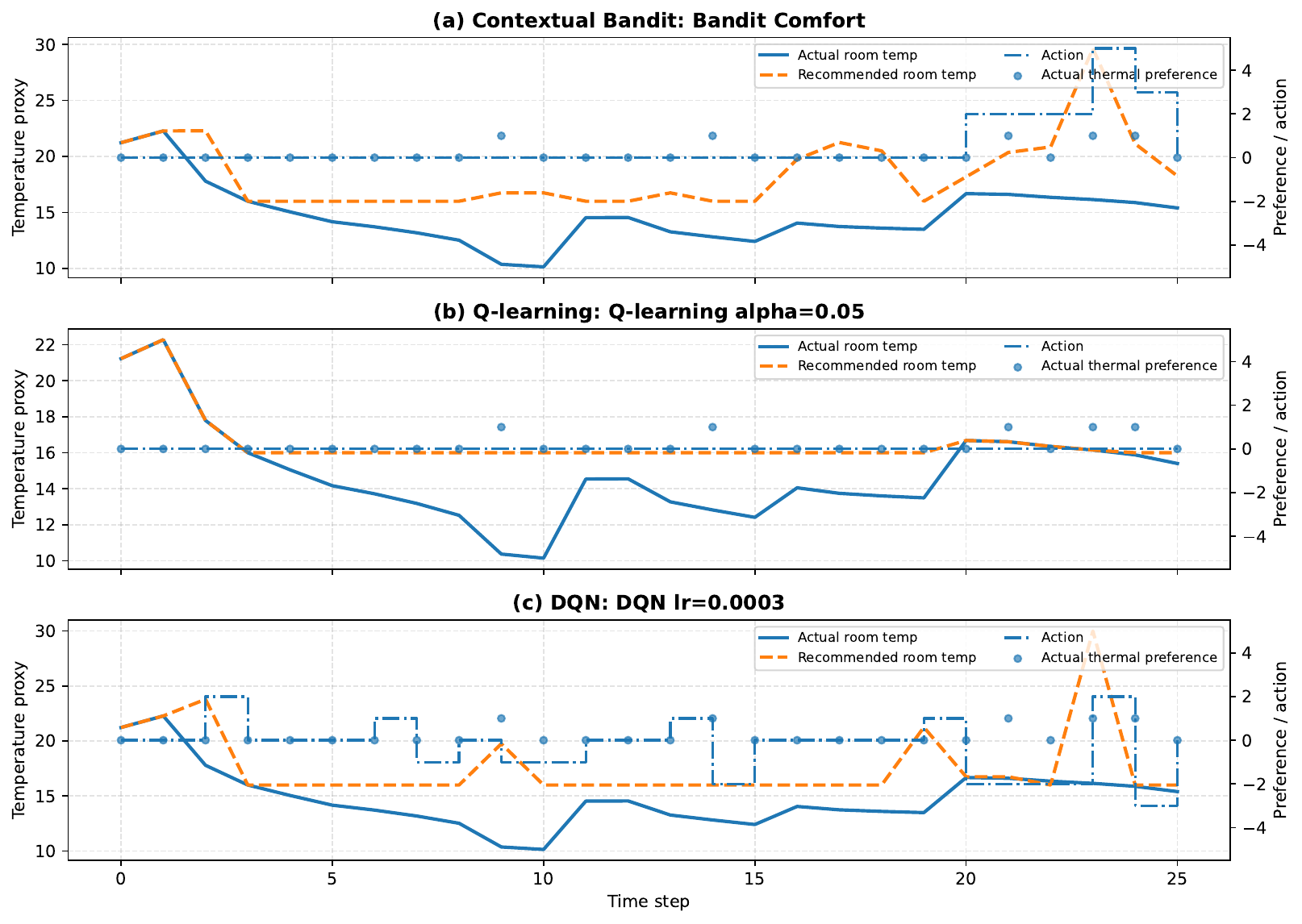}
        \caption{Subject 9: Representative closed-loop thermal regulation using Phys+480}
        \label{fig:subject7}
    \end{subfigure}
    \caption{Comparison of representative closed-loop proxy thermal-policy behaviour across two subjects under Phys+480. In the plotted legends, ``Actual room temp'' denotes the observed temperature proxy derived from the dataset, while ``Recommended room temp'' denotes the controller-generated equivalent environmental-temperature proxy. These trajectories do not represent measured room-temperature regulation or physical HVAC actuation.}
    \label{fig:subject_comparison1}
\end{figure*}

\begin{figure*}[t]
    \centering
    \begin{subfigure}[t]{0.49\textwidth}
        \centering
        \includegraphics[width=\textwidth]{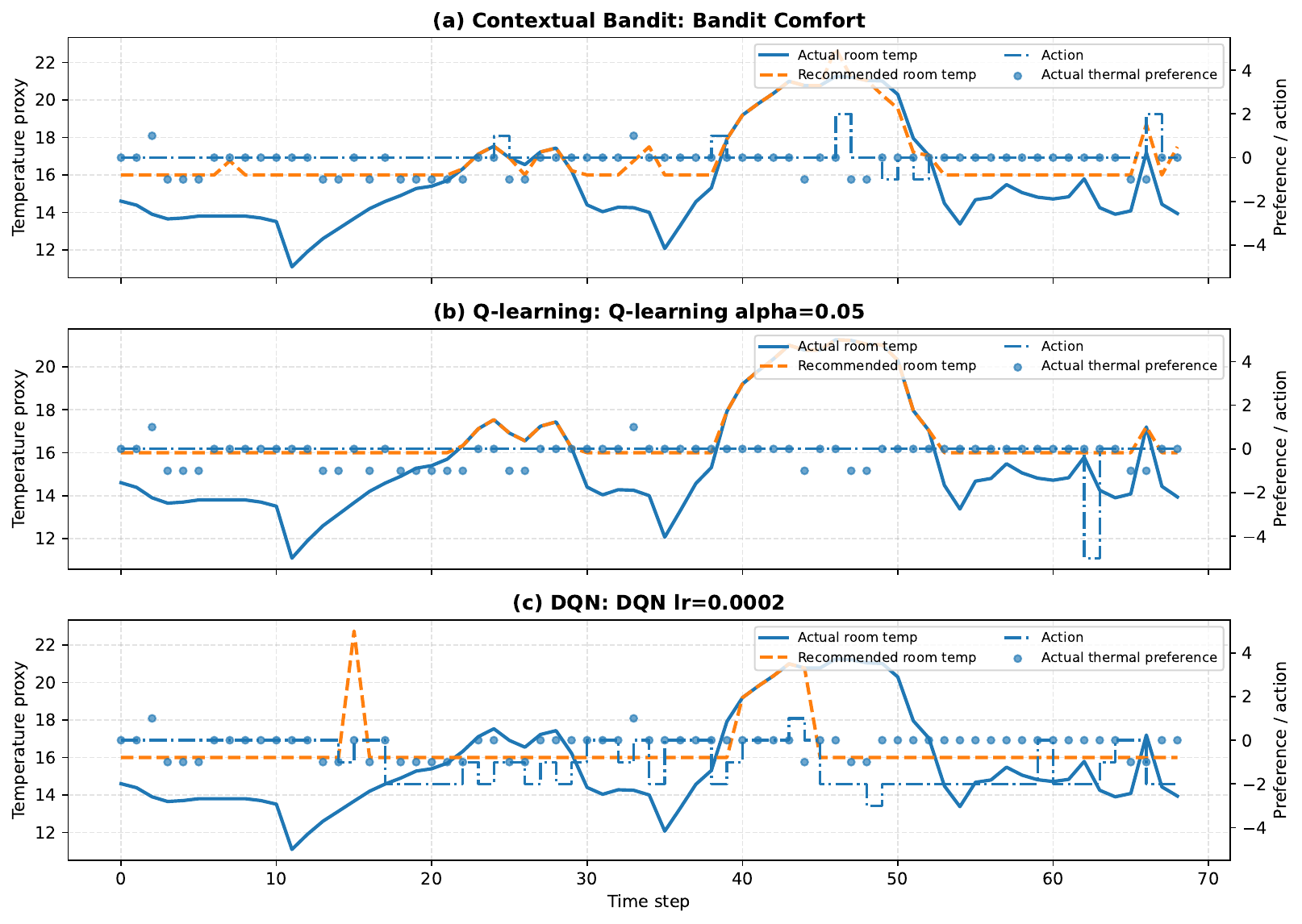}
        \caption{Subject 7: Representative closed-loop thermal regulation using Phys+60}
        \label{fig:subject9}
    \end{subfigure}
    \hfill
    \begin{subfigure}[t]{0.49\textwidth}
        \centering
        \includegraphics[width=\textwidth]{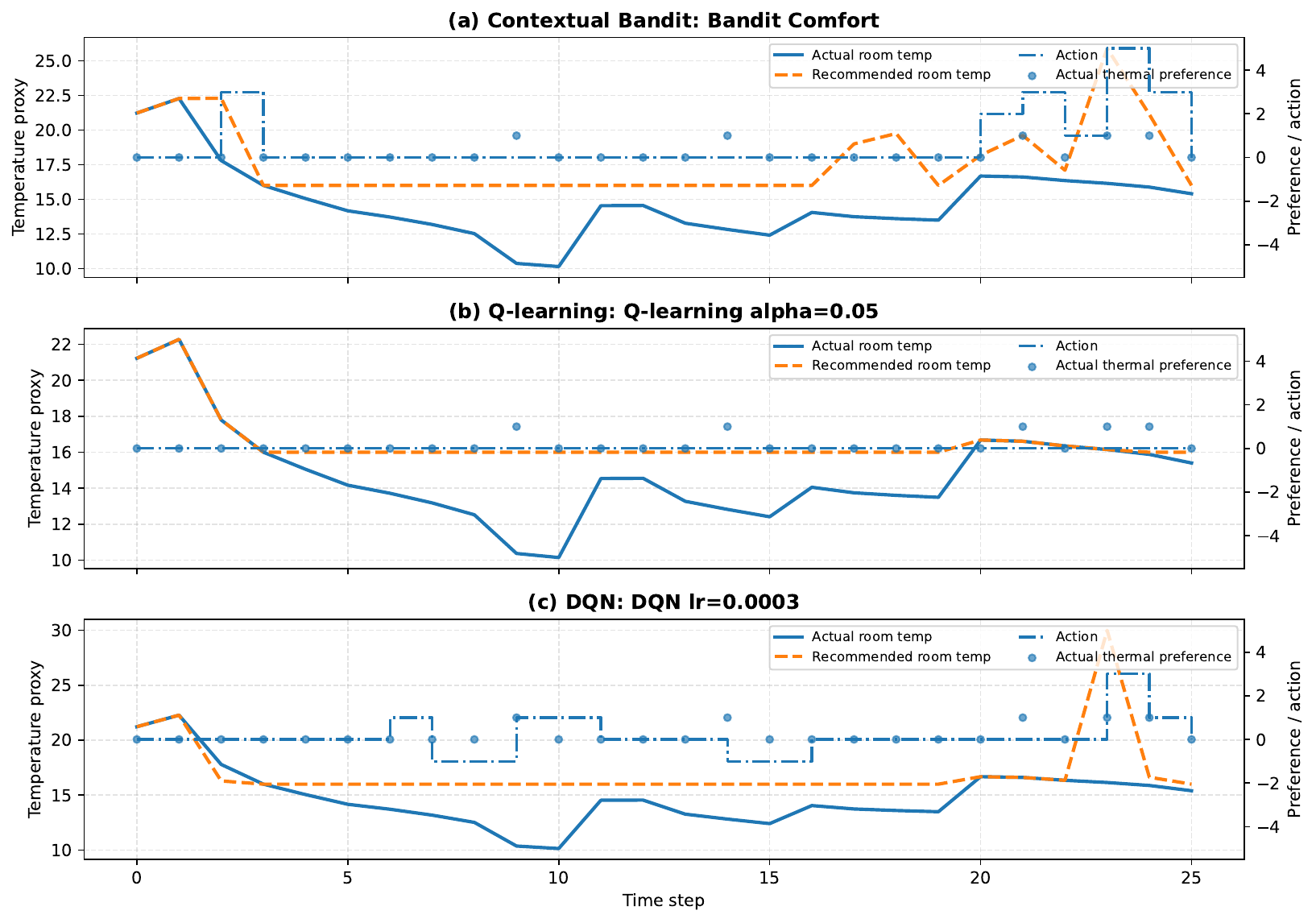}
        \caption{Subject 9: Representative closed-loop thermal regulation using Phys+60}
        \label{fig:subject7}
    \end{subfigure}
    \caption{Comparison of representative closed-loop proxy thermal-policy behaviour across two subjects under Phys+60. In the plotted legends, ``Actual room temp'' denotes the observed temperature proxy derived from the dataset, while ``Recommended room temp'' denotes the controller-generated equivalent environmental-temperature proxy. These trajectories do not represent measured room-temperature regulation or physical HVAC actuation.}
    \label{fig:subject_comparison2}
\end{figure*}

\begin{figure*}[t]
    \centering
    \begin{subfigure}[t]{0.49\textwidth}
        \centering
        \includegraphics[width=\textwidth]{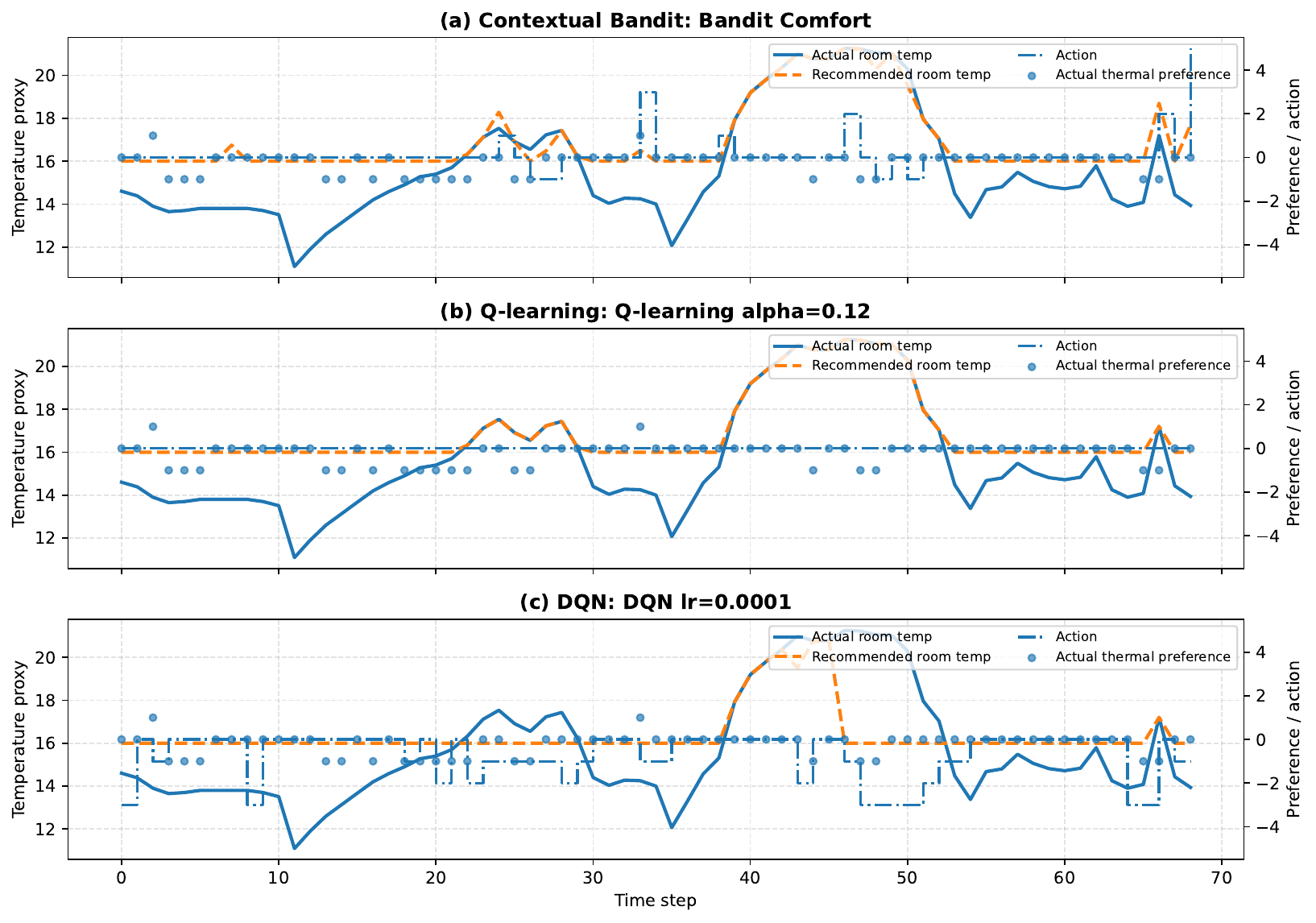}
        \caption{Subject 7: Representative closed-loop thermal regulation using Phys+60+480}
        \label{fig:subject9}
    \end{subfigure}
    \hfill
    \begin{subfigure}[t]{0.49\textwidth}
        \centering
        \includegraphics[width=\textwidth]{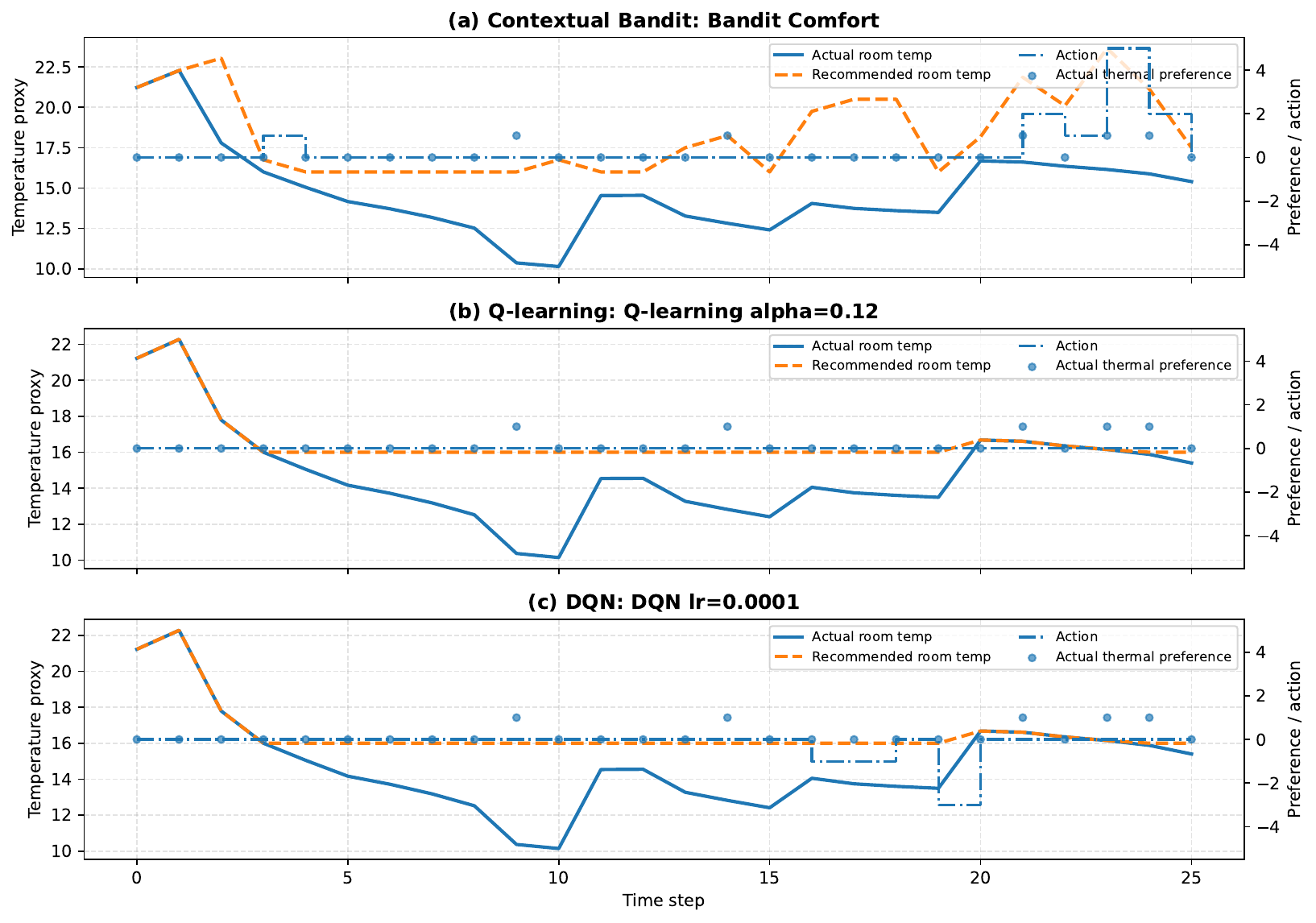}
        \caption{Subject 9: Representative closed-loop thermal regulation using Phys+60+480}
        \label{fig:subject7}
    \end{subfigure}
     \caption{Comparison of representative closed-loop proxy thermal-policy behaviour across two subjects under Phys+60+480. In the plotted legends, ``Actual room temp'' denotes the observed temperature proxy derived from the dataset, while ``Recommended room temp'' denotes the controller-generated equivalent environmental-temperature proxy. These trajectories do not represent measured room-temperature regulation or physical HVAC actuation.}
    \label{fig:subject_comparison3}
\end{figure*}

Figures~\ref{fig:subject_comparison1} - \ref{fig:subject_comparison3} illustrate representative closed-loop regulation behaviour for Subjects~7 and~9 across the three oracle feature representations. The trajectories show how the selected \gls*{cb}, \gls*{ql} and \gls*{dqn} controllers convert Comfort Oracle predictions into temperature-control actions over time, showing the trade-off between comfort maximisation and intervention efficiency. For consistency with the original visualisation outputs, the figure legends retain the labels ``Actual room temp'' and ``Recommended room temp''; throughout this section, these terms refer to the observed and controller-generated temperature proxies defined in Section~\ref{sec:stage2}, rather than measured indoor temperatures. 

Across the examples, \gls*{cb} produced the most responsive regulation behaviour, with larger and more frequent temperature adjustments. This explains its higher comfort probability and reward in Table~\ref{tab:stage2_controller_performance}, but also its lower reward-per-energy values compared with \gls*{ql}. In contrast, \gls*{ql} produced smoother trajectories with fewer interventions and a dominant ``No Change'' action profile, resulting in the highest reward-per-energy and comfort-per-energy values. \gls*{dqn} showed a more variable control pattern, with more frequent switching and larger action changes than \gls*{ql}, indicating greater policy flexibility but reduced energy efficiency. The three feature representations produced broadly similar regulation patterns, supporting the Stage~1 finding that controller policy had a stronger influence than the environmental feature window. The differences between Subjects~7 and~9 further show that the approach adapts control behaviour to participant-specific thermal preference dynamics rather than applying a fixed temperature-control strategy.

\subsubsection{Action Agreement and Control Behaviour} \label{sec:stage2_action_behaviour}
Table~\ref{tab:stage2_action_agreement} evaluates how closely the selected Stage~2 controller actions agree with the recorded occupant thermal preferences. This analysis complements Table~\ref{tab:stage2_controller_performance}. While Table~\ref{tab:stage2_controller_performance} assesses optimisation performance in terms of comfort, reward and efficiency, Table~\ref{tab:stage2_action_agreement} evaluates behavioural alignment with the observed preference labels.
\begin{table*}
\centering
\scriptsize
\caption{Action agreement performance of the selected Stage~2 controllers. Only the best variant from each controller family, as selected in Table~\ref{tab:stage2_controller_performance}, is reported.}
\label{tab:stage2_action_agreement}
\begin{tabular}{lllcccc}
\toprule
Oracle & Controller & Best Variant & Accuracy & Balanced Accuracy & Macro F1 & Cohen's kappa \\
\midrule
Phys+60 & CB & Bandit Comfort & 0.622 & 0.369 & 0.367 & 0.043 \\
Phys+60 & QL & QL alpha=0.16 & 0.691 & 0.332 & 0.285 & -0.012 \\
Phys+60 & DQN & DQN lr=0.0002 & 0.442 & 0.458 & 0.395 & 0.129 \\
Phys+480 & CB & Bandit Comfort & 0.604 & 0.368 & 0.374 & 0.045 \\
Phys+480 & QL & QL alpha=0.16 & 0.688 & 0.327 & 0.278 & -0.021 \\
Phys+480 & DQN & DQN lr=0.0001 & 0.412 & 0.392 & 0.328 & 0.052 \\
Phys+60+480 & CB & Bandit Comfort & 0.613 & 0.372 & 0.371 & 0.051 \\
Phys+60+480 & QL & QL alpha=0.16 & 0.691 & 0.332 & 0.285 & -0.012 \\
Phys+60+480 & DQN & DQN lr=0.0001 & 0.384 & 0.410 & 0.342 & 0.088 \\
\bottomrule
\end{tabular}
\end{table*}
\gls*{ql} achieved the highest overall action accuracy across all feature representations, with accuracies of 0.691 for Phys+60, 0.688 for Phys+480 and 0.691 for Phys+60+480. However, its balanced accuracy remained low, ranging from 0.327 to 0.332, while Macro F$_1$ remained between 0.278 and 0.285 and Cohen's $\kappa$ was negative across all feature sets. This indicates that the high accuracy was largely driven by frequent selection of the dominant ``No Change'' action rather than balanced recognition of cooler, no-change and warmer preferences. This behaviour is consistent with the conservative \gls*{ql} policy reported in Table~\ref{tab:stage2_controller_performance}, where more than 97\% of its actions are ``No Change''. \gls*{cb} achieved lower overall accuracy than QL, ranging from 0.604 to 0.622, but produced more consistent behavioural agreement, with balanced accuracy between 0.368 and 0.372, Macro F$_1$ between 0.367 and 0.374, and positive Cohen's $\kappa$ values across all feature representations. This suggests that \gls*{cb} was more responsive to non-neutral preference states than \gls*{ql}, although it remained primarily reward-driven. \gls*{dqn} showed the strongest minority-class sensitivity, particularly under Phys+60, where it achieved the highest balanced accuracy (0.458), Macro F$_1$ (0.395) and Cohen's $\kappa$ (0.129), despite a lower overall accuracy of 0.442. This confirms that DQN is less biased toward the dominant class but produced more variable control behaviour.

\subsubsection{Statistical Comparison of Controllers}
Table~\ref{tab:stage2_friedman_tests} presents the Friedman statistical comparison of the three reinforcement learning controller families across the participant cohort. Unlike the Stage~1 analysis, where no statistically significant differences are observed between either the candidate learning algorithms or feature representations, the Stage~2 results demonstrate that controller selection has a significant influence on closed-loop thermal regulation.

Across all three oracle feature representations, statistically significant differences are observed between CB, QL and DQN for comfort, cumulative reward, reward-per-energy, HVAC intervention magnitude and action switching rate (all $p<0.001$). These findings confirm that the observed differences in controller behaviour are consistent across participants and are not attributable to random variation. In particular, while the three personalised Comfort Oracle feature representations produced broadly comparable predictive performance in Stage~1, the reinforcement learning policy exerted a substantially greater influence on thermal regulation performance than the underlying feature representation.
\begin{table*}
\centering
\scriptsize
\caption{Friedman statistical comparison of the selected Stage~2 controllers across participants for each oracle feature representation.}
\label{tab:stage2_friedman_tests}
\begin{tabular}{lllccc}
\toprule
Oracle & Metric & Test & N & Statistic & $p$-value \\
\midrule
Phys+60 & Comfort & Friedman & 14 & 18.43 & 0.0000996 \\
Phys+60 & Reward & Friedman & 14 & 22.29 & 0.0000145 \\
Phys+60 & Reward/Energy & Friedman & 14 & 15.86 & 0.0003603 \\
Phys+60 & HVAC $\Delta T$  & Friedman & 14 & 18.43 & 0.0000996 \\
Phys+60 & Action switch rate & Friedman & 14 & 19.13 & 0.0000701 \\
Phys+480 & Comfort & Friedman & 14 & 20.76 & 0.0000310 \\
Phys+480 & Reward & Friedman & 14 & 21.57 & 0.0000207 \\
Phys+480 & Reward/Energy & Friedman & 14 & 15.43 & 0.0004464 \\
Phys+480 & HVAC $\Delta T$  & Friedman & 14 & 15.96 & 0.0003416 \\
Phys+480 & Action switch rate & Friedman & 14 & 18.73 & 0.0000856 \\
Phys+60+480 & Comfort & Friedman & 14 & 16.62 & 0.0002463 \\
Phys+60+480 & Reward & Friedman & 14 & 21.57 & 0.0000207 \\
Phys+60+480 & Reward/Energy & Friedman & 14 & 16.71 & 0.0002347 \\
Phys+60+480 & HVAC $\Delta T$  & Friedman & 14 & 19.11 & 0.0000708 \\
Phys+60+480 & Action switch rate & Friedman & 14 & 22.27 & 0.0000146 \\
\bottomrule
\end{tabular}
\end{table*}
These statistical results support the behavioural observations reported in Sections~\ref{sec:stage2_controller_performance} and~\ref{sec:stage2_tradeoff}. CB consistently prioritised occupant comfort and cumulative reward through more responsive temperature adjustments, QL adopted a conservative policy that maximised energy efficiency with minimal interventions, and DQN provided a more adaptive but less stable control strategy. The significant differences in HVAC intervention magnitude and action switching rate further demonstrate that the three controllers learn fundamentally different regulation policies despite operating on the same personalised Comfort Oracle outputs.

Consequently, pairwise Wilcoxon signed-rank tests with Holm correction are performed to identify the specific controller pairs responsible for these differences. A combined Stage~2 statistical analysis demonstrates that, once personalised thermal preference prediction has been established, the reinforcement learning strategy becomes the dominant factor governing adaptive thermal comfort control. This extends previous personalised thermal comfort studies, including \cite{liu2019personal}, by showing that personalised prediction alone is insufficient; the downstream control policy plays an equally important role in determining comfort, intervention behaviour and operational efficiency.

\subsection{Discussion} \label{sec:Discussion} The proposed two-stage approach demonstrates that personalised thermal comfort prediction and reinforcement learning-based intervention-policy generation provide complementary capabilities for intelligent thermal decision support. Stage~1 establishes participant-specific Comfort Oracles capable of capturing individual thermal preference patterns, while Stage~2 shows how these predictions can be translated into sequential proxy interventions with different comfort, intervention-efficiency and policy- stability characteristics. The discussion is organised around three dimensions of the proposed approach: (i) the importance of personalisation for representing occupant-specific thermal behaviour, (ii) the role of reinforcement learning in transforming personalised predictions into adaptive control policies, and (iii) the implications of the proposed approach for future occupant-centric smart building systems.

\subsubsection{Personalisation as the Foundation for Adaptive Thermal Control}
A key finding of this study is that personalisation forms the foundation of effective intelligent thermal control. Stage~1 demonstrated substantial inter-participant variability in thermal preference prediction, with no single learning algorithm or feature representation consistently outperforming the others across all participants. Although Phys+480 achieved the highest average predictive accuracy, the differences between feature representations and ensemble models are not statistically significant, indicating that the dominant source of variation arises from individual physiological and behavioural characteristics rather than model architecture. Consequently, selecting participant-specific Comfort Oracles provides a more appropriate representation of occupant thermal preference than adopting a single population-wide prediction model. These findings reinforce the growing body of personalised thermal comfort research that challenges the assumptions underpinning conventional PMV-based control strategies \citep{fanger1970thermal}. Rather than assuming that occupants sharing the same environmental conditions experience similar comfort, recent studies have demonstrated that physiological responses, behavioural adaptation and contextual factors vary considerably between individuals \citep{kim2018personal,kim2018personal2,liu2019personal,abdelrahman2022personal}. The participant-specific Comfort Oracles developed in this work therefore provide an adaptive representation of individual thermal preference that can accommodate this variability and serve as a reliable state representation for downstream control.

\subsubsection{From Prediction to Intelligent Decision-Making}
Although accurate personalised prediction is essential, prediction alone does not specify what environmental intervention should follow. The primary contribution of this paper is therefore the integration of
personalised Comfort Oracles with reinforcement learning to generate and compare sequential intervention policies. Unlike previous personalised comfort studies, which primarily terminate at prediction, and our earlier physiologically controlled HVAC framework, which relied on deterministic consensus-based temperature adjustment, the proposed approach closes the decision loop through reinforcement learning, enabling adaptive intervention policies that balance personalised comfort and intervention efficiency \cite{liu2019personal, ihianle2022towards}. The Stage~2 results demonstrate that, once personalised thermal preference has been established, controller design becomes the dominant factor governing adaptive regulation. This contrasts with Stage~1, where only modest differences are observed between feature representations. The \gls*{cb}s consistently maximised comfort probability and cumulative reward through more responsive temperature adjustments, whereas \gls*{ql} learned a conservative policy that minimised interventions while achieving the highest reward-per-energy and comfort-per-energy ratios. \gls*{dqn} demonstrates the best policy flexibility and improved recognition of minority thermal preference transitions, but its increased action variability results in lower overall control efficiency. The statistical analysis confirms that these behavioural differences are significant across all evaluated metrics, demonstrating that \gls*{rl} policy has a substantially greater influence on closed-loop performance than the underlying feature representation. These observations are consistent with \gls*{rl} theory, where different optimisation mechanisms naturally produce different control policies. \gls*{cb}s optimise immediate expected reward, \gls*{ql} seeks long-term value through incremental policy improvement, while \gls*{dqn} approximates the action-value function using deep neural networks, enabling greater representational capacity but potentially introducing higher policy variance \citep{mnih2015human}. Therefore, controller selection should reflect operational priorities, whether maximising occupant comfort, reducing intervention magnitude or balancing both objectives.

\subsubsection{Implications for Occupant-Centric Smart Buildings}
The proposed approach provides a practical pathway towards occupant-centric HVAC systems capable of continuously adapting environmental conditions to individual physiological responses. Rather than regulating indoor temperature using fixed comfort assumptions, the approach combines wearable sensing, personalised comfort modelling and reinforcement learning to generate adaptive environmental-intervention policies that evolve with the occupant. In the context of this study, these policies are evaluated through equivalent environmental-temperature proxy actions representing the direction and relative magnitude of prospective HVAC intervention. This approach aligns with the broader vision of intelligent and human-centric buildings, where environmental control responds dynamically to occupant behaviour rather than predefined schedules or static comfort models. The accompanying Streamlit-based proof-of-concept implementation further demonstrates how personalised thermal comfort prediction and reinforcement learning can be visualised and evaluated interactively before deployment within operational building management systems \footnote{Web App proof-of-concept of the system is available online at: \url{https://huggingface.co/spaces/Isibor/Thermal_Comfort_Controller}}. This provides an accessible platform for analysing controller behaviour, comparing reinforcement learning strategies and supporting future digital commissioning of occupant-centric HVAC systems.

Several limitations should nevertheless be acknowledged. Stage~2 was evaluated within a proxy control environment derived from the available wearable dataset rather than through a physical HVAC system, calibrated building simulator or validated zone-temperature model. The dataset does not contain directly measured indoor air temperature, thermostat setpoints, actuator dynamics, HVAC operating states or electrical power consumption. Accordingly, the equivalent environmental-temperature actions should be interpreted as proxy control signals that represent the direction and relative magnitude of prospective intervention, rather than realised thermostat commands or measured indoor-temperature changes. Similarly, the reported reward-efficiency and comfort-efficiency metrics quantify performance relative to intervention magnitude and should not be interpreted as direct evidence of building energy efficiency or absolute energy savings. The results therefore demonstrate the comparative behaviour of personalised reinforcement learning policies within a common proxy environment, rather than physically validated HVAC regulation. Nevertheless, this approach establishes a reproducible methodology for integrating personalised thermal comfort prediction with sequential intervention-policy generation and provides a strong foundation for future validation using instrumented buildings, calibrated digital twins, measured HVAC operation and real-time energy data.

\section{Conclusion}
\label{sec:conclusion}
This paper presented a two-stage personalised thermal comfort approach that integrates wearable physiological sensing, participant-specific Comfort Oracles and reinforcement learning for adaptive thermal
intervention-policy generation. Unlike existing personalised thermal comfort approaches that primarily focus on prediction, the proposed approach combines personalised thermal preference modelling with sequential decision-making, enabling closed and human-in-the-loop environmental regulation that continuously adapts to individual occupant needs.

Experimental evaluation using the Liu et al., \cite{liu2019personal} wearable thermal comfort dataset demonstrates that personalised Comfort Oracles provide a robust representation of individual thermal preference despite substantial inter-participant variability. While only modest differences are observed between the evaluated feature representations and ensemble learning models during Stage~1, Stage~2 showed that \gls*{rl} policy has a much greater influence on adaptive thermal regulation. \gls*{cb}s consistently maximised occupant comfort and cumulative reward, \gls*{ql} achieved the highest reward and comfort efficiency through a more conservative control strategy, and \gls*{dqn} demonstrated the best policy flexibility but with increased intervention variability. Statistical analysis confirms significant differences between the controller \gls*{rl} models across all evaluated control metrics, highlighting the importance of controller selection once reliable personalised thermal preference prediction has been established.

The proposed approach demonstrates the feasibility of coupling participant-specific Comfort Oracles with reinforcement learning to generate personalised sequential intervention policies within a proxy control environment. The findings establish comparative policy behaviour in terms of predicted comfort, reward, intervention magnitude and action stability; they do not constitute physical HVAC validation, measured indoor-temperature regulation or evidence of building energy savings. Future work will therefore validate this approach using instrumented buildings and calibrated digital twins with measured indoor conditions, HVAC operation and energy consumption, while extending the approach to continuous-action and multi-occupant control settings.

\printcredits

%\bibliographystyle{cas-model2-names}

%\bibliography{cas-refs}

\bibliographystyle{IEEEtranN}
\bibliography{cas-refs}

\appendix
\section*{Appendix A. Additional Tables\label{sec:appendixA}}

\setcounter{table}{0}
\renewcommand{\thetable}{A\thesection.\arabic{table}}

\onecolumn

\begin{table}[h!]
\centering
\scriptsize
\setlength{\tabcolsep}{3pt}
\renewcommand{\arraystretch}{1.0}

\caption{Participant-level Stage~1 performance across all feature representations. Values are Accuracy/Macro F$_1$/Cohen's $\kappa$.}
\label{tab:stage1_full_all_feature_sets}

\begin{subtable}{\textwidth}
\centering
\caption{Physiology}
\label{tab:stage1_full_physiology_acc_f1_kappa}
\begin{tabular}{lcccccccccccccc}
\toprule
\textbf{Model} & \textbf{1} & \textbf{2} & \textbf{3} & \textbf{4} & \textbf{5} & \textbf{6} & \textbf{7} & \textbf{8} & \textbf{9} & \textbf{10} & \textbf{11} & \textbf{12} & \textbf{13} & \textbf{14} \\
\midrule
RF & \makecell{0.65/\\0.43/\\0.30} & \makecell{0.49/\\0.40/\\0.01} & \makecell{0.63/\\0.39/\\0.13} & \makecell{0.93/\\0.48/\\0.00} & \makecell{0.85/\\0.47/\\0.38} & \makecell{0.45/\\0.36/\\0.09} & \makecell{0.66/\\0.31/\\-0.03} & \makecell{0.89/\\0.31/\\0.00} & \makecell{0.80/\\0.36/\\0.09} & \makecell{0.65/\\0.38/\\0.08} & \makecell{0.84/\\0.47/\\0.17} & \makecell{0.59/\\0.59/\\0.18} & \makecell{0.71/\\0.63/\\0.26} & \makecell{0.83/\\0.30/\\0.00} \\
\midrule
GB & \makecell{0.68/\\0.45/\\0.36} & \makecell{0.43/\\0.43/\\0.04} & \makecell{0.60/\\0.37/\\0.08} & \makecell{0.91/\\0.32/\\-0.01} & \makecell{0.85/\\0.47/\\0.38} & \makecell{0.47/\\0.37/\\0.11} & \makecell{0.66/\\0.31/\\-0.02} & \makecell{0.89/\\0.40/\\0.17} & \makecell{0.80/\\0.37/\\0.11} & \makecell{0.67/\\0.43/\\0.16} & \makecell{0.84/\\0.47/\\0.17} & \makecell{0.65/\\0.65/\\0.30} & \makecell{0.63/\\0.57/\\0.14} & \makecell{0.83/\\0.36/\\0.11} \\
\midrule
ET & \makecell{0.58/\\0.41/\\0.24} & \makecell{0.39/\\0.36/\\-0.05} & \makecell{0.62/\\0.38/\\0.11} & \makecell{0.89/\\0.31/\\-0.02} & \makecell{0.85/\\0.50/\\0.44} & \makecell{0.43/\\0.37/\\0.07} & \makecell{0.61/\\0.44/\\0.06} & \makecell{0.89/\\0.54/\\0.17} & \makecell{0.80/\\0.36/\\0.10} & \makecell{0.62/\\0.36/\\0.09} & \makecell{0.79/\\0.42/\\0.11} & \makecell{0.62/\\0.42/\\0.27} & \makecell{0.73/\\0.65/\\0.31} & \makecell{0.82/\\0.30/\\-0.03} \\
\bottomrule
\end{tabular}
\end{subtable}

\begin{subtable}{\textwidth}
\centering
\caption{Physiology + Env60}
\label{tab:stage1_full_physiology_env60_acc_f1_kappa}
\begin{tabular}{lcccccccccccccc}
\toprule
\textbf{Model} & \textbf{1} & \textbf{2} & \textbf{3} & \textbf{4} & \textbf{5} & \textbf{6} & \textbf{7} & \textbf{8} & \textbf{9} & \textbf{10} & \textbf{11} & \textbf{12} & \textbf{13} & \textbf{14} \\
\midrule
RF & \makecell{0.55/\\0.37/\\0.14} & \makecell{0.57/\\0.38/\\0.03} & \makecell{0.65/\\0.40/\\0.15} & \makecell{0.87/\\0.47/\\-0.07} & \makecell{0.82/\\0.37/\\0.14} & \makecell{0.59/\\0.39/\\0.18} & \makecell{0.77/\\0.38/\\0.20} & \makecell{0.89/\\0.31/\\0.00} & \makecell{0.80/\\0.36/\\0.09} & \makecell{0.75/\\0.50/\\0.25} & \makecell{0.85/\\0.31/\\-0.05} & \makecell{0.53/\\0.50/\\0.09} & \makecell{0.68/\\0.55/\\0.14} & \makecell{0.83/\\0.30/\\0.00} \\
\midrule
GB & \makecell{0.65/\\0.45/\\0.36} & \makecell{0.55/\\0.47/\\0.13} & \makecell{0.69/\\0.51/\\0.31} & \makecell{0.91/\\0.48/\\-0.03} & \makecell{0.82/\\0.42/\\0.23} & \makecell{0.39/\\0.27/\\0.07} & \makecell{0.68/\\0.35/\\0.08} & \makecell{0.89/\\0.31/\\0.00} & \makecell{0.76/\\0.35/\\0.06} & \makecell{0.65/\\0.38/\\0.11} & \makecell{0.83/\\0.30/\\-0.06} & \makecell{0.62/\\0.61/\\0.25} & \makecell{0.66/\\0.59/\\0.18} & \makecell{0.83/\\0.36/\\0.11} \\
\midrule
ET & \makecell{0.58/\\0.39/\\0.17} & \makecell{0.49/\\0.37/\\-0.00} & \makecell{0.66/\\0.45/\\0.22} & \makecell{0.78/\\0.44/\\-0.11} & \makecell{0.85/\\0.50/\\0.44} & \makecell{0.45/\\0.32/\\0.06} & \makecell{0.71/\\0.48/\\0.13} & \makecell{0.89/\\0.31/\\0.00} & \makecell{0.78/\\0.35/\\0.06} & \makecell{0.63/\\0.33/\\-0.03} & \makecell{0.85/\\0.40/\\0.09} & \makecell{0.59/\\0.56/\\0.20} & \makecell{0.68/\\0.52/\\0.10} & \makecell{0.83/\\0.30/\\0.00} \\
\bottomrule
\end{tabular}
\end{subtable}

\begin{subtable}{\textwidth}
\centering
\scriptsize
\caption{Physiology + Env60 + Env480}
\label{tab:stage1_full_physiology_env60_env480_acc_f1_kappa}
\begin{tabular}{lcccccccccccccc}
\toprule
\textbf{Model} & \textbf{1} & \textbf{2} & \textbf{3} & \textbf{4} & \textbf{5} & \textbf{6} & \textbf{7} & \textbf{8} & \textbf{9} & \textbf{10} & \textbf{11} & \textbf{12} & \textbf{13} & \textbf{14} \\
\midrule
RF & \makecell{0.55/\\0.33/\\0.08} & \makecell{0.59/\\0.39/\\0.01} & \makecell{0.62/\\0.30/\\0.05} & \makecell{0.82/\\0.45/\\-0.10} & \makecell{0.84/\\0.42/\\0.27} & \makecell{0.53/\\0.33/\\0.06} & \makecell{0.76/\\0.37/\\0.17} & \makecell{0.89/\\0.31/\\0.00} & \makecell{0.82/\\0.41/\\0.23} & \makecell{0.60/\\0.26/\\-0.05} & \makecell{0.86/\\0.31/\\-0.04} & \makecell{0.59/\\0.55/\\0.21} & \makecell{0.66/\\0.50/\\0.06} & \makecell{0.83/\\0.30/\\0.00} \\
\midrule
GB & \makecell{0.58/\\0.34/\\0.11} & \makecell{0.49/\\0.41/\\-0.00} & \makecell{0.69/\\0.49/\\0.29} & \makecell{0.89/\\0.31/\\-0.04} & \makecell{0.80/\\0.42/\\0.20} & \makecell{0.41/\\0.29/\\0.10} & \makecell{0.70/\\0.37/\\0.11} & \makecell{0.87/\\0.31/\\-0.02} & \makecell{0.76/\\0.35/\\0.05} & \makecell{0.62/\\0.31/\\-0.02} & \makecell{0.85/\\0.39/\\0.09} & \makecell{0.65/\\0.64/\\0.31} & \makecell{0.66/\\0.59/\\0.18} & \makecell{0.83/\\0.36/\\0.11} \\
\midrule
ET & \makecell{0.55/\\0.36/\\0.08} & \makecell{0.57/\\0.41/\\0.05} & \makecell{0.68/\\0.46/\\0.25} & \makecell{0.78/\\0.51/\\0.04} & \makecell{0.85/\\0.47/\\0.38} & \makecell{0.41/\\0.29/\\0.06} & \makecell{0.75/\\0.35/\\0.14} & \makecell{0.89/\\0.31/\\0.00} & \makecell{0.76/\\0.38/\\0.12} & \makecell{0.62/\\0.34/\\-0.00} & \makecell{0.86/\\0.31/\\-0.04} & \makecell{0.62/\\0.60/\\0.26} & \makecell{0.66/\\0.46/\\0.01} & \makecell{0.82/\\0.30/\\-0.03} \\
\bottomrule
\end{tabular}
\end{subtable}

\end{table}

\begin{table}
\centering
\scriptsize
\caption{Stage~2 controller performance across all oracle feature representations and controller variants.}
\label{tab:stage2_all_controller_variants}
\begin{tabular}{lllcccccccc}
\toprule
Oracle & Controller & Variant & Comfort & Reward & HVAC $\Delta T$ & Reward/Energy & Comfort/Energy &  Decrease &  No Change &  Increase \\
\midrule
Phys+60 & CB & Bandit Comfort & 0.700 & 89.942 & 3.342 & 26.917 & 0.210 & 0.093 & 0.817 & 0.090 \\
Phys+60 & CB & Bandit Balanced & 0.698 & 89.344 & 3.202 & 27.900 & 0.218 & 0.087 & 0.838 & 0.075 \\
Phys+60 & CB & Bandit Energy & 0.695 & 88.703 & 3.100 & 28.616 & 0.224 & 0.081 & 0.855 & 0.064 \\
Phys+60 & QL & QL alpha=0.16 & 0.668 & 84.442 & 2.628 & 32.132 & 0.254 & 0.019 & 0.978 & 0.003 \\
Phys+60 & QL & QL alpha=0.08 & 0.667 & 84.314 & 2.712 & 31.087 & 0.246 & 0.031 & 0.960 & 0.009 \\
Phys+60 & QL & QL alpha=0.05 & 0.667 & 84.112 & 2.708 & 31.059 & 0.246 & 0.031 & 0.954 & 0.015 \\
Phys+60 & QL & QL alpha=0.12 & 0.667 & 83.804 & 2.750 & 30.469 & 0.242 & 0.046 & 0.953 & 0.001 \\
Phys+60 & DQN & DQN lr=0.0002 & 0.678 & 84.382 & 3.343 & 25.238 & 0.203 & 0.417 & 0.386 & 0.198 \\
Phys+60 & DQN & DQN lr=0.0005 & 0.678 & 83.932 & 3.650 & 22.996 & 0.186 & 0.538 & 0.260 & 0.202 \\
Phys+60 & DQN & DQN lr=0.0003 & 0.679 & 83.703 & 3.426 & 24.433 & 0.198 & 0.551 & 0.270 & 0.179 \\
Phys+60 & DQN & DQN lr=0.0001 & 0.678 & 83.533 & 3.798 & 21.994 & 0.178 & 0.592 & 0.309 & 0.099 \\
Phys+480 & CB & Bandit Comfort & 0.684 & 87.417 & 3.602 & 24.269 & 0.190 & 0.084 & 0.781 & 0.134 \\
Phys+480 & CB  & Bandit Balanced & 0.682 & 86.923 & 3.379 & 25.726 & 0.202 & 0.075 & 0.808 & 0.117 \\
Phys+480 & CB  & Bandit Energy & 0.678 & 86.326 & 3.184 & 27.110 & 0.213 & 0.058 & 0.848 & 0.095 \\
Phys+480 & QL & QL alpha=0.16 & 0.658 & 82.764 & 2.642 & 31.324 & 0.249 & 0.019 & 0.978 & 0.003 \\
Phys+480 & QL & QL alpha=0.08 & 0.658 & 82.553 & 2.707 & 30.494 & 0.243 & 0.034 & 0.963 & 0.003 \\
Phys+480 & QL & QL alpha=0.05 & 0.659 & 82.552 & 2.697 & 30.604 & 0.244 & 0.034 & 0.956 & 0.010 \\
Phys+480 & QL & QL alpha=0.12 & 0.658 & 82.263 & 2.756 & 29.849 & 0.239 & 0.046 & 0.953 & 0.001 \\
Phys+480 & DQN & DQN lr=0.0001 & 0.664 & 82.374 & 3.225 & 25.542 & 0.206 & 0.477 & 0.427 & 0.096 \\
Phys+480 & DQN & DQN lr=0.0002 & 0.666 & 82.258 & 3.377 & 24.357 & 0.197 & 0.403 & 0.440 & 0.157 \\
Phys+480 & DQN & DQN lr=0.0005 & 0.664 & 82.078 & 3.436 & 23.887 & 0.193 & 0.419 & 0.377 & 0.204 \\
Phys+480 & DQN & DQN lr=0.0003 & 0.664 & 81.781 & 3.310 & 24.704 & 0.201 & 0.446 & 0.443 & 0.111 \\
Phys+60+480 & CB  & Bandit Comfort & 0.696 & 87.382 & 3.548 & 24.627 & 0.196 & 0.103 & 0.795 & 0.102 \\
Phys+60+480 & CB & Bandit Balanced & 0.694 & 86.917 & 3.346 & 25.972 & 0.207 & 0.089 & 0.824 & 0.087 \\
Phys+60+480 & CB  & Bandit Energy & 0.691 & 86.357 & 3.175 & 27.200 & 0.218 & 0.071 & 0.857 & 0.072 \\
Phys+60+480 & QL & QL alpha=0.16 & 0.668 & 82.764 & 2.639 & 31.359 & 0.253 & 0.019 & 0.978 & 0.003 \\
Phys+60+480 & QL & QL alpha=0.05 & 0.668 & 82.492 & 2.724 & 30.289 & 0.245 & 0.035 & 0.950 & 0.015 \\
Phys+60+480 & QL & QL alpha=0.08 & 0.667 & 82.340 & 2.773 & 29.689 & 0.241 & 0.040 & 0.951 & 0.009 \\
Phys+60+480 & QL & QL alpha=0.12 & 0.668 & 82.224 & 2.750 & 29.895 & 0.243 & 0.044 & 0.954 & 0.001 \\
Phys+60+480 & DQN & DQN lr=0.0001 & 0.677 & 82.809 & 3.568 & 23.206 & 0.190 & 0.541 & 0.329 & 0.130 \\
Phys+60+480 & DQN & DQN lr=0.0002 & 0.679 & 82.638 & 3.292 & 25.105 & 0.206 & 0.470 & 0.375 & 0.155 \\
Phys+60+480 & DQN & DQN lr=0.0005 & 0.674 & 82.306 & 3.205 & 25.677 & 0.210 & 0.464 & 0.362 & 0.174 \\
Phys+60+480 & DQN & DQN lr=0.0003 & 0.677 & 82.203 & 3.364 & 24.435 & 0.201 & 0.474 & 0.409 & 0.117 \\
\bottomrule
\end{tabular}
\end{table}

\end{document}